\documentclass[lettersize,journal]{IEEEtran}
\usepackage{amsmath,amsfonts}
\usepackage{algorithmic}
\usepackage{algorithm}
\usepackage{array}
\usepackage[caption=false,font=normalsize,labelfont=sf,textfont=sf]{subfig}
\usepackage{textcomp}
\usepackage{stfloats}
\usepackage{url}
\usepackage{verbatim}
\usepackage{graphicx}
\usepackage{cite}
\usepackage{multicol}
\usepackage{multirow}
\usepackage{booktabs}
\usepackage{setspace}
\usepackage{tabularx}
\usepackage{makecell}
\usepackage{hyperref} 
\usepackage[rgb]{xcolor}
\definecolor{deepgreen}{rgb}{0,0.5,0}
\begin{document}

\title{OE-BevSeg: An Object Informed and Environment Aware Multimodal Framework for Bird's-eye-view Vehicle Semantic Segmentation}

\author{Jian Sun, Yuqi Dai, Chi-Man Vong, Qing Xu, Shengbo Eben Li, Jianqiang Wang, Lei He*, Keqiang Li

\thanks{
This study is supported by National Natural Science Foundation of China, Science Fund for Creative Research Groups (Grant No.52221005)

Jian Sun, Yuqi Dai, Qing Xu, Shengbo Eben Li, Jianqiang Wang, Lei He and Keqiang Li are with the School of Vehicle and Mobility, Tsinghua University, Beijing, China, and also with the State Key Laboratory of Intelligent Green Vehicle and Mobility, Tsinghua University, Beijing, China (e-mail: daiyuqi@mail.tsinghua.edu.cn; qingxu@tsinghua.edu.cn; lishbo@tsinghua.edu.cn; helei2023@tsinghua.edu.cn; likq@tsinghua.edu.cn)
\textit{(Corresponding author: Lei He.)}

Jian Sun and Chi-Man Vong are with the Department of Computer and Information Science, University of Macau, Macau, China, (e-mail: sjian1015@163.com; cmvong@umac.mo)

}
}



\maketitle

\begin{abstract}

Bird's-eye-view (BEV) semantic segmentation is becoming crucial in autonomous driving systems. It realizes ego-vehicle surrounding environment perception by projecting 2D multi-view images into 3D world space. Recently, BEV segmentation has made notable progress, attributed to better view transformation modules, larger image encoders, or more temporal information. However, there are still two issues:  1) a lack of effective understanding and enhancement of BEV space features, particularly in accurately capturing long-distance environmental features and 2) recognizing fine details of target objects. To address these issues, we propose OE-BevSeg, an end-to-end multimodal framework that enhances BEV segmentation performance through global environment-aware perception and local target object enhancement. OE-BevSeg employs an environment-aware BEV compressor. Based on prior knowledge about the main composition of the BEV surrounding environment varying with the increase of distance intervals, long-sequence global modeling is utilized to improve the model’s understanding and perception of the environment. From the perspective of enriching target object information in segmentation results, we introduce the center-informed object enhancement module, using centerness information to supervise and guide the segmentation head, thereby enhancing segmentation performance from a local enhancement perspective. Additionally, we designed a multimodal fusion branch that integrates multi-view RGB image features with radar/LiDAR features, achieving significant performance improvements. Extensive experiments show that, whether in camera-only or multimodal fusion BEV segmentation tasks, our approach achieves state-of-the-art results by a large margin on the nuScenes dataset for vehicle segmentation, demonstrating superior applicability in the field of autonomous driving. Our code will be released at \href{https://github.com/SunJ1025/OE-BevSeg}{https://github.com/SunJ1025/OE-BevSeg}.

\end{abstract}


\begin{figure}
\centering
\includegraphics[width=3.5in]{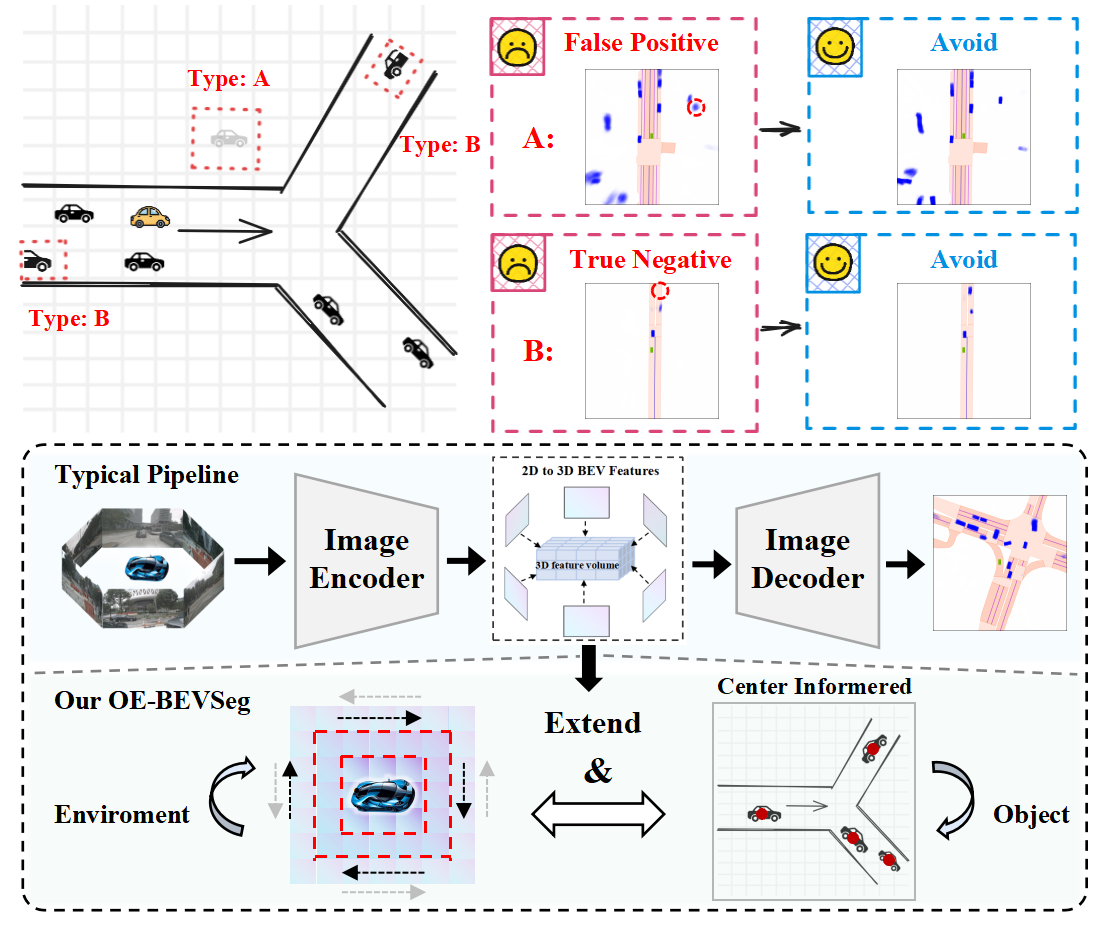}
\caption{The pipline of our proposed OE-BevSeg. The BEV feature is enhanced in terms of both environment and target object perspectives to improve the performance of segmentation.
}
\label{fig_1}
\end{figure}

\section{Introduction}

\IEEEPARstart{R}{ecent} years, BEV perception has played an increasingly important role in the field of autonomous driving \cite{teng2023motion,he2022diff,he2024neural,han20234d}. Almost all mainstream autonomous driving systems, robotics, etc.,  have adopted the BEV paradigm for 3D space perception. Due to the low cost of cameras, converting perspective view (PV) to BEV view \cite{hu2021fiery,li2022bevformer,zhou2022cross, xu2024surround } using multiple cameras has attracted extensive attention. As a pixel-level classification algorithm, semantic segmentation \cite{sun2022bi,harley2023simple,liu2023bird} predicts high-precision semantic perception labels for autonomous vehicles. BEV semantic segmentation's powerful representation of the surrounding environment provides rich contextual and geometric information for autonomous driving planning and decision. Meanwhile, the BEV space facilitates multi-modal fusion by merging data from different sensors (e.g., cameras, radar, LiDAR) into a unified space. BEVFusion\cite{liu2023bevfusion} proposes a method for fusing RGB images and LiDAR data. It employs two parallel networks to separately extract image features and point cloud features, and finally performs fusion in the BEV space. BEVFusion \cite{liu2023bevfusion} has demonstrated that incorporating LiDAR data can significantly enhance the accuracy and robustness of 3D object detection in autonomous driving. In this work, inspired by \cite{harley2023simple}, we integrate radar/LiDAR information by simply rasterizing the radar/LiDAR data and combining it with camera features. Experimental results demonstrate that the utilization of radar or LiDAR significantly enhances the performance BEV segmentation by a large margin. 

Generally speaking, BEV perception algorithms consist of three important fundamental components, which are PV image encoder, perspective transformation module, and decoder. Most pipelines mainly focus on improving these three aspects. Specifically, this involves handle the view discrepancy between the surrounding environment and a fixed-size BEV space, with the aim of achieving pixel-level semantic segmentation of different vehicle targets. 

Existing methods typically involve designing a distinctive viewpoint transformation module between the camera image space and the BEV space. Early methods are mainly geometric projection-based approaches. Inverse perspective mapping (IPM) \cite{bertozz1998stereo} is a representative algorithm among them, which requires using the camera's intrinsic and extrinsic parameters or camera pose for transformation. However, this method is based on the assumption that the ground is flat and is sensitive to vehicle motion. Deep learning-based methods, such as Multi-Layer Perceptron (MLP) and Transformer \cite{vaswani2017attention}, have demonstrated improved robustness. In this work, we do not use the computationally expensive Multi-scale Deformable Attention \cite{zhu2020deformable}. Instead, we employ a simple bilinear sampling strategy \cite{harley2023simple} to achieve efficient perspective transformation.

However, after completing the perspective transformation from PV to BEV, directly feeding the BEV features into the decoder for simple upsampling does not fully exploit the potential of the BEV feature. This approach may lose pivotal cues in the BEV space, resulting in less robust performance in some challenging scenarios. BEVDet \cite{huang2021bevdet} proposed a BEV encoder to further encodes the BEV feature. They utilized the residual structure of ResNet to perceive scale, direction, and velocity within the BEV space. UIF-BEV \cite{ren2024uif} introduced camera underlying information to achieve interaction between temporal and spatial data in the BEV space. OCBEV\cite{qi2024ocbev} utilizes high-confidence locations as decoder queries, adding heatmap supervision to the BEV feature. The above methods further process the BEV feature using convolutional networks or transformer architectures, leading to improved segmentation performance. However, they lack consideration of the long-distance hierarchical environment of the BEV space. In recent advancements in semantic segmentation, Mamba \cite{gu2023mamba} architecture has improved the computational efficiency of the SSM model \cite{gu2021efficiently}, demonstrating excellent performance in long-sequence modeling and global information dependency. 

In this paper, we present OE-BevSeg, which adopts the classic Encoder-Decoder structure and introduces a fusion branch to fuse RGB image and point cloud data. We optimize the network from two aspects: environmental perception and target object enhancement. From the perspective of environmental perception, we introduce Mamba into BEV semantic segmentation. Unlike existing models, our method leverages Mamba's efficient long-sequence modeling capability, achieving linear computational complexity while maintaining global enviroment awareness. Compared to Transformers of similar scale, Mamba has five times the throughput \cite{gu2023mamba}. 

We explored the environmental characteristics of BEV space and proposed that the urban environmental elements in BEV space can be divided into three hierarchical stages as the distance from the ego-vehicle increases: road, buildings, and sky. Based on this, we proposed an improved Mamba scan mechanism. Inspired by \cite{qi2024ocbev}, in addition to environment-aware perception, object-centric modeling is key to BEV segmentation. Therefore, we fully utilized centerness information \cite{hu2021fiery} to enrich target area information, enhancing the model's robustness in challenging scenarios.

In summary, our contributions are as follows:

\begin{itemize}

\item We present an end-to-end multimodal framework for the BEV vehicle semantic segmentation. To the best of our knowledge, this is the first work to specifically design Mamba global modeling for the BEV segmentation.

\item We propose the Environment-aware BEV Compressor, which leverages Mamba's global receptive field to enhance BEV feature's ability to perceive and understand the long-distance surrounding environment.

\item We improve Mamba's scanning mechanism in BEV space. Based on the enviroment prior knowledge, we propose a Bi-Surround Scan approach that prioritizes the modeling of relevant hierarchical environmental components, enabling model to capture the spatial relationships of BEV features and enhance the model's environment perception capability, thereby better distinguishing between foreground and background.

\item From an object-centric modeling perspective, we propose the Center-Informed Object Enhancement module, which uses centerness information to supervise and guide the segmentation head. Spatial attention and multi-view deformable cross-attention are used to enhance the model's focus on local target object areas, enriching the segmentation results with target object-related information.

\end{itemize}

\begin{figure*}
\centering
\includegraphics[width=7in]{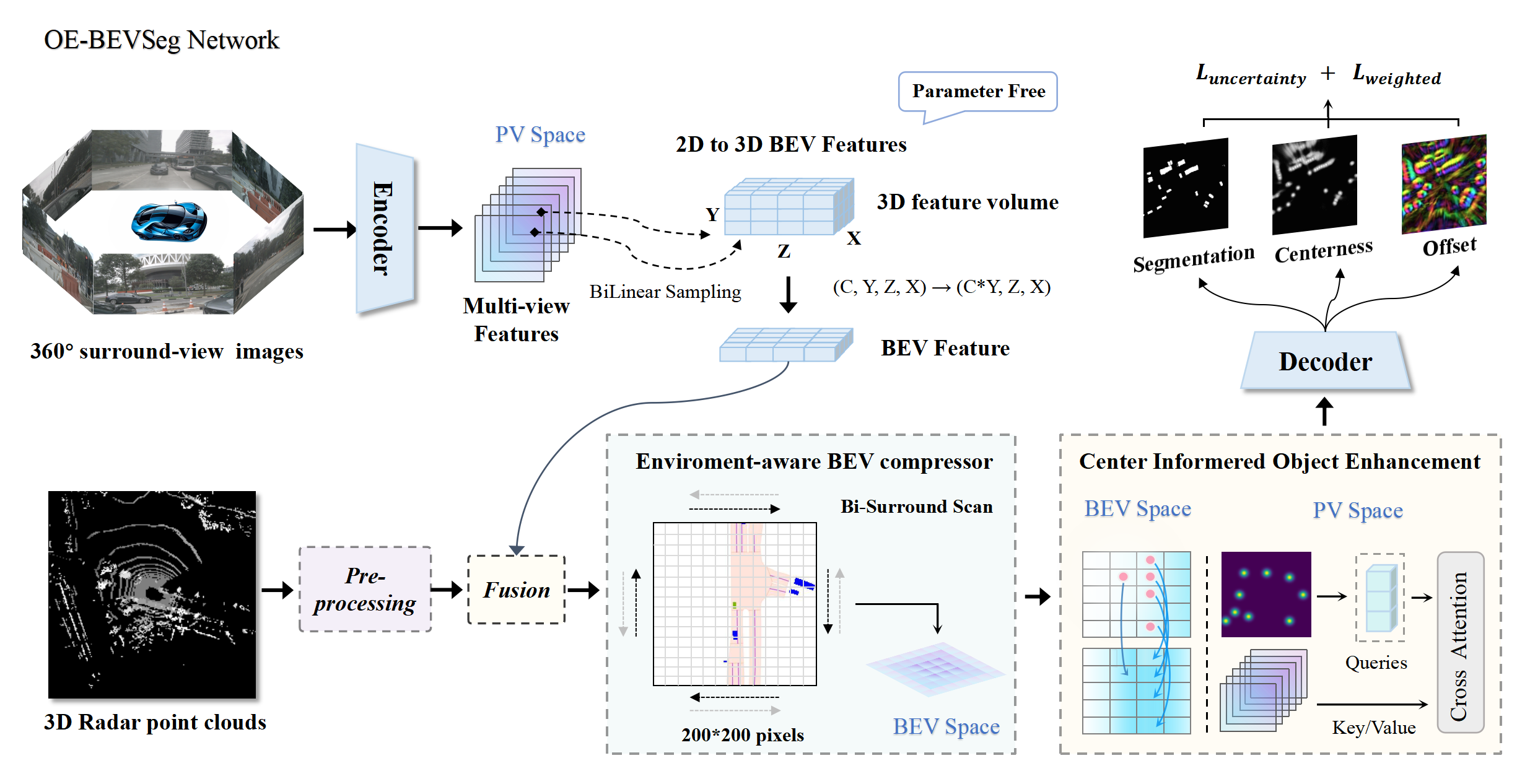}
\caption{The pipline of our proposed OE-BevSeg. The input consists of two parts: surround-view RGB images $I_{s}$ and radar/LIDAR point clouds $P$. The encoder extracts multi-view features $\left \{ F_{p,k} \right \} _{k=1}^{K}$ in the PV space. Through parameter-free perspective transformation, $\left \{ F_{p,k} \right \} _{k=1}^{K}$ are lifted from 2D features to 3D features $F_{3d}$. The preprocessed point cloud features $F_{p}$ are then fused with $F_{3d}$ for multimodal fusion. The fused features $F_{f}$ are processed through our designed EBC and CIOE modules. We use a multi-task head to handle the Decoder's output. Besides segmentation, our model also predicts centerness and offset for regularization.
}
\label{fig_2}
\end{figure*}

\section{Related work}
This paper contributes to the state-of-the-art BEV semantic segmentation algorithm. In this section, we will introduce related works on bird's-eye-view (BEV) semantic segmentation algorithms, including camera-only methods and multimodal approaches. Specifically, we will review State Space-related models, which are introduced into the BEV segmentation field for the first time in this work.
\subsection{BEV Segmentation in Intelligent Traffic Systems.}
In recent years, bird's eye view (BEV) perception algorithms have made a significant contribution in the fields of intelligent transportation and autonomous driving. 
BEV semantic segmentation is critical as it provides essential local maps in today's intelligent traffic systems, where perception is prioritized over High-definition (HD) mapping. 

{\bf Camerea-based BEV Perception Methods.} 
VPN \cite{pan2020cross} is one of the pioneering approaches in cross-view semantic segmentation. It introduced the View Parsing Network in simulated 3D environments, utilizing multiple MLP layers to regress the transformation matrix from PV to BEV. Based on VPN, PON \cite{roddick2020predicting} introduced a semantic Bayesian occupancy grid to establish map relationships across multiple frames. It also utilized a feature pyramid to extract multi-resolution features. CVT \cite{zhou2022cross} is a Transformer-based method that utilizes cross-attention to perform the conversion from PV space to BEV space. BEVFormer \cite{li2022bevformer} and BEVSegFormer \cite{peng2023bevsegformer} leverage deformable attention in a DETR-style \cite{zhu2020deformable} approach to enhance view transformation and feature representation. BEVFormer makes full use of spatiotemporal information by introducing a spatial cross-attention and temporal self-attention to aggregate spatial and temporal features. BEVSegFormer, by contrast, focuses on flexibly handling BEV segmentation tasks under arbitrary (single or multiple) cameras without requiring intrinsic and extrinsic camera parameters.

{\bf Multimodal Perception Methods.} Cameras and radars are two of the most common sensors in the field of autonomous driving. Radar point cloud features typically contain information that complements camera data. Fishing Net \cite{hendy2020fishing} employs the same MLP as VPN for view conversion and extends BEV segmentation through posterior multimodal fusion using radar and LiDAR features. CRN \cite{kim2023crn} utilizes a well-designed multimodal fusion module and proposes a two-stage fusion method, which employs deformable attention to integrate camera and radar data effectively. In contrast to CRN, simpleBEV \cite{harley2023simple} proposes a lifting method that does not rely on depth estimation. It highlights that batch size and input resolution significantly impact BEV segmentation performance. The study also demonstrates that using a simple method to concatenate camera and rasterized radar data can effectively enhance segmentation performance.

\subsection{Developement of State Space Model.}
State Space Model (SSM) have shown promising performance in long sequence modeling for its linear-time inference, parallelizable training, and robust performance. Unlike CNNs and Transformers, State Space Model (SSM) is inspired by traditional control theory, which maps input sequences to state representations. Structured State Space for Sequences (S4) \cite{gu2021efficiently} effectively handles discretized data by utilizing the zero-order hold method. The use of Highest Polynomial Powered Operator (HiPPO) \cite{gu2020hippo} initialization significantly enhances the long-range modeling capabilities of the S4 model. However, S4 cannot adaptively adjust based on the input, meaning it cannot selectively focus on important parts of the input. Mamba incorporates selective information processing, enabling dynamic handling of input data. This allows the model to selectively remember or ignore different parts of the input, thereby improving the utilization of historical information. 

Recent works, VMamba \cite{liu2024vmamba} and Vim \cite{zhu2024vision}, introduce the Mamba model to computer vision tasks. VMamba proposes a Cross-Scan Module to conduct selective scan in four different directions, which allows the model to maintain a global receptive field while achieving linear computational complexity. Vim introduces a Bidirectional SSM that performs global modeling in both forward and backward directions. \cite{zhao2024rs,ma2024u,xing2024segmamba}, demonstrate the effectiveness of the Mamba model in medical and remote sensing image segmentation. Inspired by these works, we introduce Mamba into BEV semantic segmentation task to enhance the model's global perception capabilities to the long-distance surrounding enviroments of the BEV space.

\begin{figure*}
\centering
\includegraphics[width=7in]{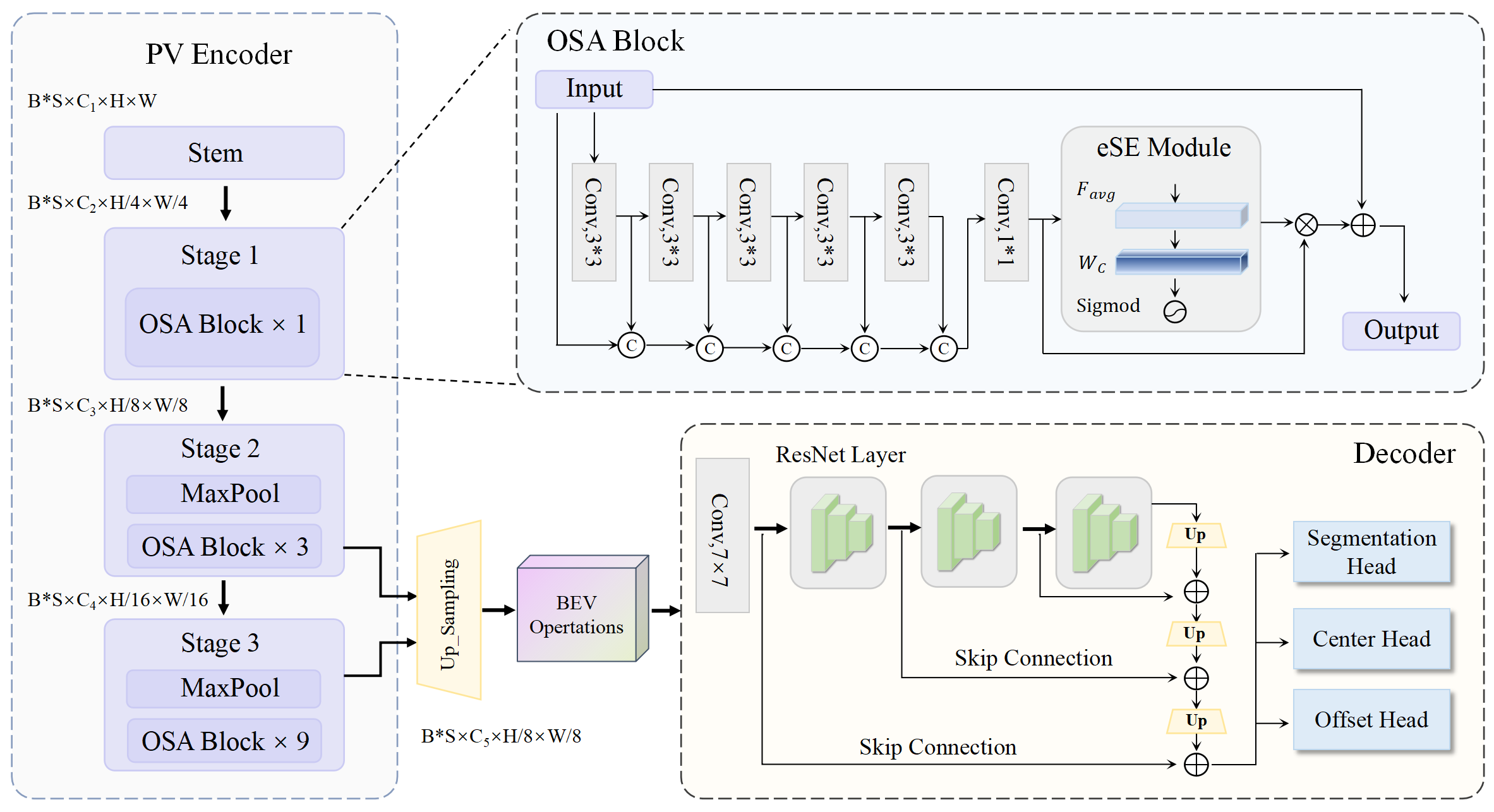}
\caption{The overall network backbone of our proposed OE-BevSeg, which can be mainly divided into two parts: the Encoder and the Decoder. The Encoder part is primarily composed of OSA blocks, while the Decoder part is mainly composed of ResNet layers. After the Encoder, a series of BEV operations are performed to complete the perspective transformation.}
\label{fig_3}
\end{figure*}

\section{Methodology}
In this work, we propose the OE-BEVSeg model, which adopts an Encoder-Decoder framework and introduces a fusion branch to integrate radar/LiDAR and RGB PV features, achieving multimodal segmentation. We utilize a simple, parameter-free BEV perspective transformation module to obtain 3D voxels through bilinear sampling. Our BEV segmentation network is optimized from two perspectives: environment perception and target object enhancement. From the perspective of environment perception, we propose the Environment-aware BEV Compressor: (1) For the BEV features aggregated from surround-view, we utilize the long-range modeling capability of the Mamba to obtain global perception features. (2) Using prior knowledge, we design a bi-surround scan method based on the characteristics of BEV perception to enhance the model's understanding of BEV space, thereby acquiring more robust BEV features. From the target object perspective, we propose the Center-Informed Object Enhancement module: (1) In the PV space, we use the centerness predicted by the center head as a query, leveraging multi-view deformable cross-attention to enhance the model's ability to predict the centerness of the target object. (2) In the BEV space, we apply a spatial cross attention mechanism to the center feature to improve the focus on the object center region in the segmentation results.

\subsection{Problem Definition and Overview}
In autonomous driving task, BEV perception  involve segmenting by projecting the vehicle's surrounding environment into a bird's-eye view. As illustrated in Fig.\ref{fig_2}, the inputs typically consist of images $I_{S}= \left \{I_{c,1},I_{c,2}...,I_{c,k}\right \}, I_{S}\in \mathbb{R}^{C\times H \times W}$ from multiple cameras and radar point cloud $P \in \mathbb{R}^{N\times Z\times X}$
, where $k$ and $N$ denote the number of cameras and the channels of the radar data, respectively. $X$, $Y$, $Z$ are used to represent the left-right, up-down, and front-back axes, respectively. After the PV sapce feature $\left \{ F_{p,k} \right \} _{k=1}^{K} \in \mathbb{R}^{d\times h\times w}$ is extracted by the encoder, it is lifted to 3D space by a parameter free bilinear sampling method\cite{harley2023simple}. Then $F_{3D} \in \mathbb{R}^{d\times Z\times Y\times X}$ is aggregated to get the BEV feature $F_{B} \in \mathbb{R}^{d\ast Y, Z, X}$, where $d$ is the dimension of the PV sapce feature. The 3D perception range represented by $F_{B}$ is $100m \times 100m$ in the $Z$ and $X$ axes, with a corresponding resolution of $200 \times 200$. The up/down span of $Y$ is set to $10m$, with a resolution of 8. The fusion of multimodal features $F_{B}$ and $F_{P}$ is achieved in BEV space to obtain the fused features $F_{f} \in \mathbb{R}^{(d*Y+N) \times Z \times X}$. In this paper, we propose the Environment-aware BEV Compressor (EBC) and Center-Informed Object Enhancement (CIOE) modules to enhance the model's environment perception ability and enrich the detail target object information. Finally, the Decoder performs upsampling and includes a multi-task head designed for segmentation and predicting centerness and offsets, which outputs the final segmentation mask $M \in \left \{  0,1\right \} ^{1 \times H \times W}$.

\subsection{Overall Architecture}
The overall architecture of our proposed OE-BevSeg illustrated in Fig.\ref{fig_3}. To capture rich contextual information, we utilized a large backbone VOVNetV2 followed by \cite{lee2019energy} as the encoder for PV perspective images, and the first three layers of a small backbone ResNet18 \cite{he2016identity} for upsampling. The input image $\left \{  I_{in,k}\right \} _{k}^{S} \in \mathbb{R}^{B*S \times C_{1} \times H \times W}$ first passes through the stem, generating features with $1/4$ width and height of the $\left \{  I_{in,k}\right \}_{k}^{S}$. $B,S,C_{1}$ represent the batchsize, camera number, input channel. The stem is composed of three layers of $3\times 3$ convolutions with strides of $2, 1,$ and $2$. The input channels of each stage $\left \{  C_{i}\right \} _{i=0}^{3} = \left \{ 256, 512, 768 \right \}$. OSA block number of each stage $\left \{  B_{i}\right \} _{i=0}^{3} = \left \{ 1, 3, 9 \right \}$. The One-Shot Aggregation (OSA) block employs the principle of dense interaction, aggregating the outputs of all previous layers in one step at the end to capture rich multi-scale texture information. The eSE Module is a crucial component of the OSA block, similar to the SENet \cite{hu2018squeeze} structure, which can further enhance segmentation performance. The eSE Module can be represented as follows:

\begin{equation}
\label{deqn_ex1a}
\mathcal{W}_{c}=\sigma \left ( FC \left (  \mathcal{A}\left ( x \right )  \right ) \right )  
\end{equation}

\begin{equation}
\label{deqn_ex1a}
F_{out}= F_{in}+  \mathcal{W}_{c} \otimes x
\end{equation}

Where $\mathcal A$ represents average pooling, $FC$ denotes fully connected layer, and $\delta$ refers to the hsigmod operation, which specifically consists of ReLU6. $\mathcal W_{c}$ represents learnable channel weights, and $\otimes$ indicates element-wise multiplication. $F_{in}$ and $F_{out}$ represent the input features and output features of the OSA block, respectively.

In the Decoder part, we use simple ResNet layers $\left \{  L_{i}  \right \}_{i=1}^{n}$, along with bilinear upsampling $\left \{ Upsample_{i}\right \}_{i=1}^{n},n=3$ and cross-layer skip connections to map the features to the output mask. By utilizing this layer-by-layer summative skip connection method, more multi-scale detail information is preserved during the upsampling process. The upsampling process can be summarized in the following equation:

\begin{equation}
\label{deqn_ex1a}
U_{F,i} =  Upsamle_{i+1}(U_{F,i+1})+S_{F,i},i=0,1,2.
\end{equation}

where $\left \{U_{F,i}\right \}_{i=0}^{2}$ denotes the feature after upsampling in $L_{i}$, and $\left \{S_{F,i}\right \}_{i=0}^{2}$ denotes the skip connection feature map in $L_{i}$. Upsampling starts at the $L_{3}$ and proceeds sequentially forward, ultimately generating an upsampled feature map $U_{F,0}$.

\begin{figure}
\centering
\includegraphics[width=3.5in]{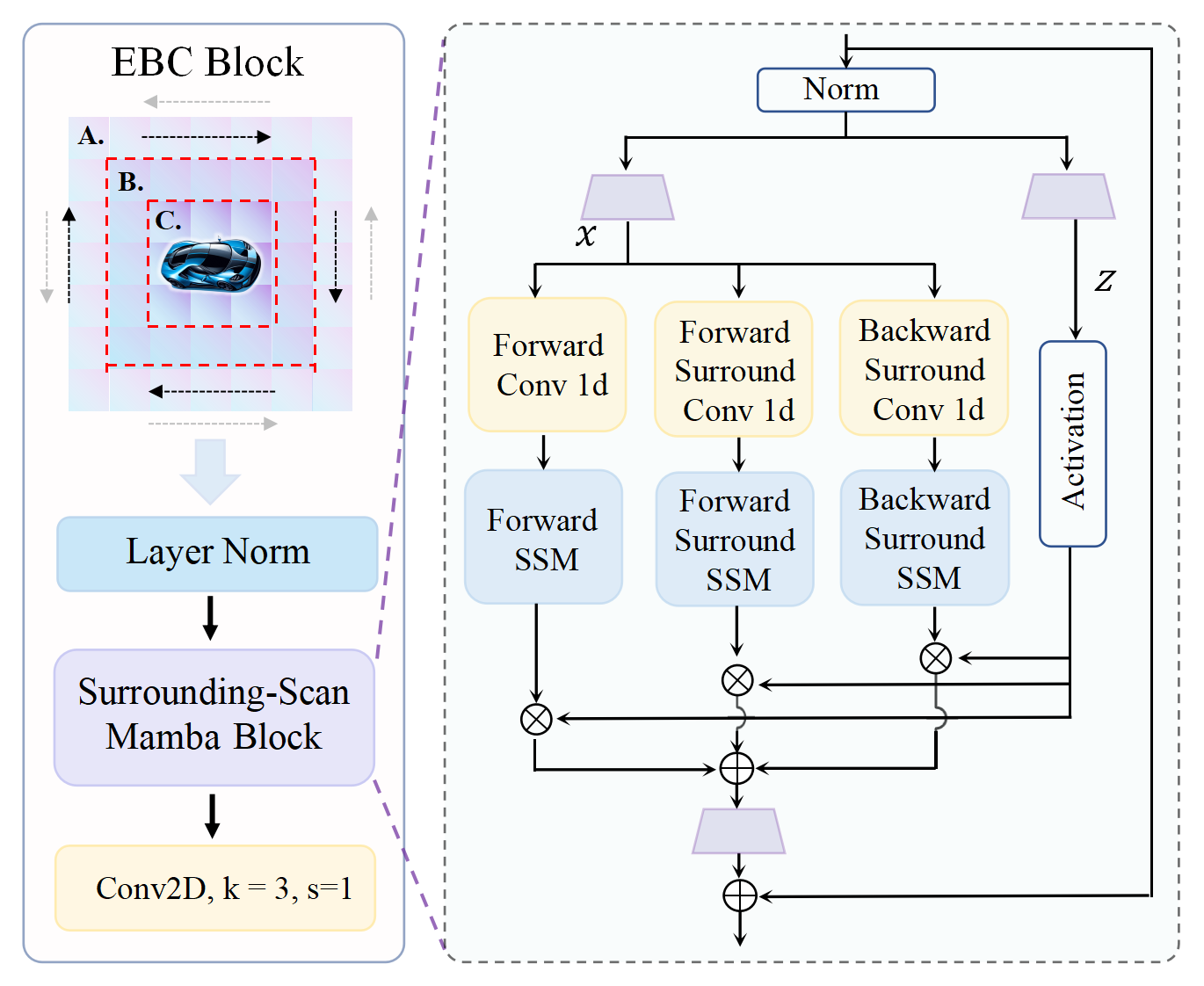}
\caption{The architecture of our proposed Environment-aware BEV Compressor (EBC). The input is BEV feature, which is processed for bi-surround scan.}
\label{fig_4}
\end{figure}

\subsection{Environment-aware BEV Compressor}
Bird’s-Eye View (BEV) features encapsulate extensive environmental semantics, which significantly influences a model’s ability to differentiate between foreground and background elements. Previous research lacks further processing of BEV features after completing the perspective transformation. Inspired by the powerful long sequences modeling and computationally efficient State Space Model (SSM), we employed the recently prominent Mamba to model the global context of the environment in the BEV space. 

With full consideration of the characteristics of the BEV perspective, we find that as the distance from the ego-vehicle increases, the surrounding environment sampled by the BEV features will change significantly. Specifically, at the closest distance to the ego-vehicle, such as $\operatorname{0-20m}$, the main component of the environment is the road surface. At a further distance of $\operatorname{20-35m}$, the main component of the environment is usually buildings. At $\operatorname{35-50m}$, the main component in the environment from multiple camera views should be the sky-related elements. Experiments in distance-phase segmentation \cite{ren2024uif,zhou2022cross} have also demonstrated this point, the segmentation results vary significantly at different distances from the ego-vehicle. Therefore, if only global features are learned while ignoring the characteristics of the driving environment, the model cannot effectively understand and perceive the BEV space. Based on the aforementioned prior knowledge, we have improved Mamba by proposing the Bi-Surround Scan mechanism to serialize the BEV features, enhancing the model's environment-aware capabilities.

\textbf{State Space Model (SSM).} State Space Model originates from control theory and represents a method of mathematically describing a problem using the minimal number of variables that fully describe the linear time-invariant system, i.e., $x(t) \in \mathbb{R} \longmapsto y(t) \in \mathbb{R}$. In continuous time $t$, the current hidden state $h_{t}\in \mathbb{R}^{N}$ is calculated using the previous hidden state $h_{t-1}\in \mathbb{R}^{N}$ and the current input $x_{t}\in \mathbb{R}$. $y_{t}\in \mathbb{R}$ is the prediction of $h_{t}\in \mathbb{R}^{N}$. Specifically, the model can be represented by the following ordinary differential equations:

\begin{equation}
\label{eq_4}
{h}'(t)=\textbf{A}h(t)+\textbf{B}x(t)
\end{equation}

\begin{equation}
\label{eq_5}
y(t)=\textbf{C}h(t)+Dx(t)
\end{equation}

Where $A \in \mathbb{C}_{N\times N}$ is the evolution parameter, $B \in \mathbb{C}^{N}$, $C \in \mathbb{C}^{N}$ and $D \in \mathbb{C}^{1}$ are the weighting parameters.

\textbf{Discretization of SSM.} 
The above SSMs cannot handle discrete data. To be applicable in deep learning, the SSMs of the continuous system need to be discretized. Structured State Space for Sequences (S4) achieves the discretization of SSMs using the zero-order hold technique, as shown in Eqn.(\ref{eq_6})-(\ref{eq_8}):

\begin{equation}
\label{eq_6}
\bar{\textbf{A}} = exp(\textbf{A}\Delta t)
\end{equation}

\begin{equation}
\label{eq_7}
\bar{\textbf{B}} =(\int_{0}^{\Delta t}  exp(\textbf{A}\Delta \tau )d\tau )\cdot \Delta \textbf{B}
\end{equation}

\begin{equation}
\label{eq_8}
\bar{\textbf{C}} =\textbf{C}
\end{equation}

$\overline{\textbf{A}}$, $\overline{\textbf{B}}$ and $\overline{\textbf{C}}$ are the discrete parameters. The discretized Eqn.(\ref{eq_4})-(\ref{eq_5}) can be transformed as:

\begin{equation}
\label{eq_9}
h(t)=\bar{\textbf{A}}h(t-1)+\bar{\textbf{B}}x(t)
\end{equation}

\begin{equation}
\label{eq_10}
y(t)=\bar{\textbf{C}}h(t)+ \bar{D}x(t)
\end{equation}

However, the parameters of $\overline{\textbf{A}}$, $\overline{\textbf{B}}$, and $\overline{\textbf{C}}$ in S4 cannot adaptively change based on different inputs, which limits the ability to effectively select important information. Mamba introduces a selection mechanism to enhance adaptability to different inputs and more effectively utilize previous information.

\textbf{Bi-Surround Scan mechanism} 

Existing studies explore new scan methods suitable for specific tasks, such as cross-scan \cite{liu2024vmamba} and backward scan \cite{zhu2024vision} to improve Mamba's image understanding capability. Inspired by the BEV space's surround sampling strategy and the typical urban environment's main components based on distance intervals, we designed a bidirectional surround scan mechanism (Bi-Surround Scan) to improve Mamba’s adaptability in processing BEV features.

In this paper, the resolution of the BEV space is $200\times 200$. The BEV features are divided into $2\times 2$ patches in the $X$ and $Z$ axes, resulting in $100\times 100$ patch embeddings. As shown in Fig.\ref{fig_4}, according to the distance intervals from the ego-vehicle, we divided the BEV feature into three parts:$\left \{ A, B, C \right \}$, corresponding to the distances of $\left \{\operatorname{0-20m}, \operatorname{20-35m}, \operatorname{35-50m} \right \}$. These three parts are hierarchically correspond to the road surface, buildings, and sky elements. Simply learning the global contextual information through serialization ignores this prior information. Therefore, based on the forward scan, we propose the forward surround scan and backward surround scan. These methods prioritize the modeling of environment-related components, allowing Mamba to learn hierarchical environmental features. The detailed process is illustrated in Algo.\ref{alg:alg1}. Given the BEV feature $F_{B}$ and the Surround-Scan Mamba block with three distinct scan branches, $S_{f}$, $S_{fs}$, and $S_{bs}$. $SiLU$ is employed as the activation function, and $z$ serves as the gating mechanism for the outputs ${y}'_{f}$, ${y}'_{fs}$, ${y}'_{bs}$. Finally, $F_{EB}$ is obtained through skip connections.

\subsection{Center-Informed Object Enhancement}
\begin{figure*}
\centering
\includegraphics[width=7in]{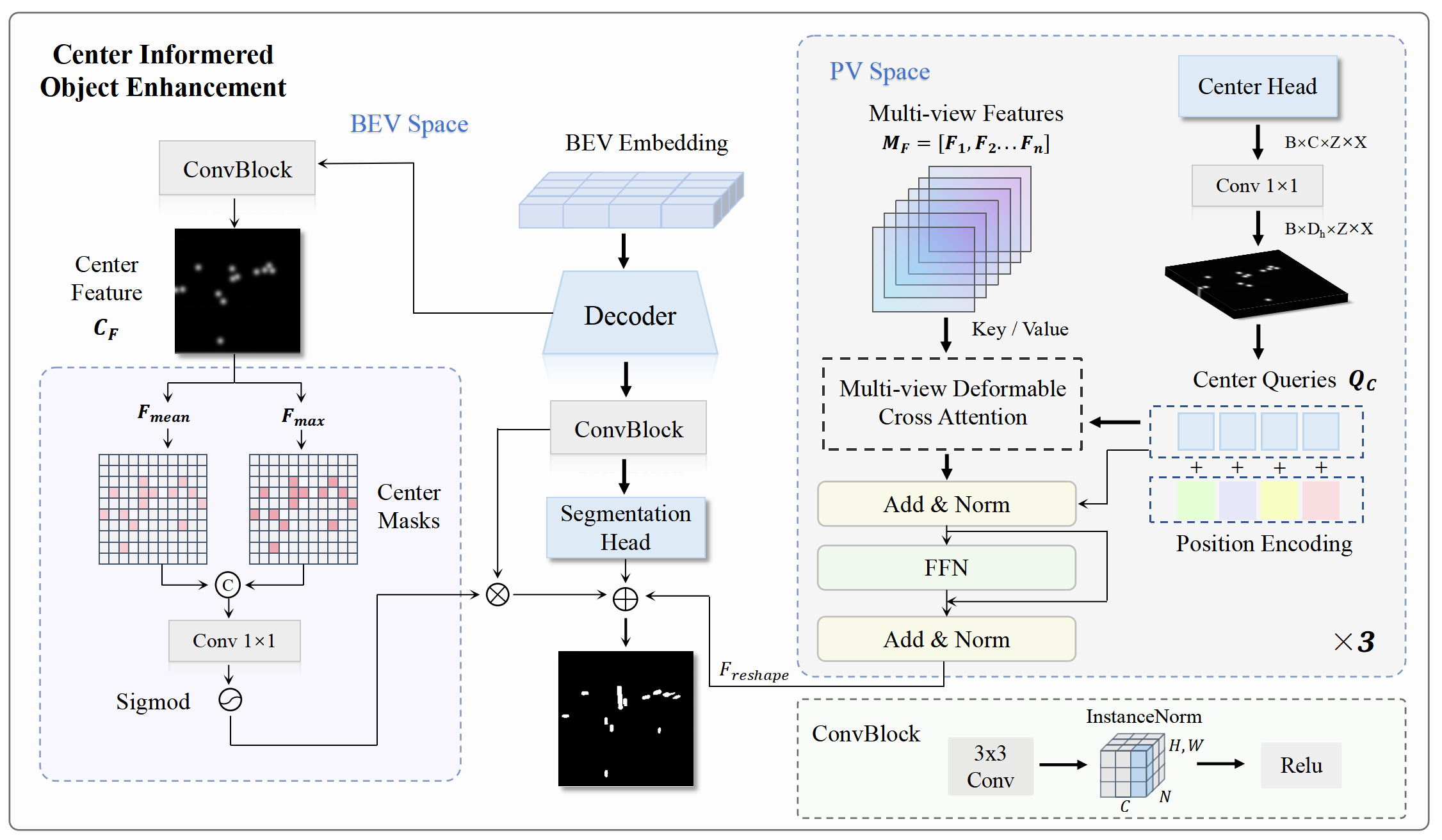}
\caption{The architecture of our proposed Center-Informed Object Enhancement (CIOE) module. We use vehicle instance centerness in both PV space and BEV space to enrich the detail target object information in the segmentation results.}
\label{fig_5}
\end{figure*}

Inspired by object-centric modeling \cite{qi2024ocbev}, we introduced a center head that uses vehicle instance centerness as supervisory information, and designed a corresponding center loss based on L2 distance. In addition to the loss function, we explicitly utilized centerness information in both PV and BEV space to enhance and improve the output of the segmentation head. As shown in Fig.\ref{fig_5}, we take the output of ConvBlock after Decoder and before center head in BEV space as center feature $C_{F}$. Taking Center Feature $C_{F}$ as input, we use spatial attention \cite{hu2018squeeze} to obtain the center masks $C_{mask}$ based on the

\begin{algorithm}[H]
\caption{Environment-aware BEV Compressor}
\label{alg:alg1}
\begin{algorithmic}
\begin{spacing}{0.2}
\STATE \textsc{\textbf{input:}} \hspace{0.1cm}  BEV feature $F_{B}:\textcolor{deepgreen}{(B,D,H,W)}$
\end{spacing}
\begin{spacing}{0.2}
\STATE {\textsc{\textbf{output:}}}\hspace{0.1cm} Enhanced BEV feature $F_{EB}:\textcolor{deepgreen}{(B,D,H,W)}$
\end{spacing}
\begin{spacing}{0.2}
\STATE {\textsc{\textbf{initialize:}}}\hspace{0.1cm}  Patch size $P=2$, Number patches $N=(\frac{H}{P}) \times (\frac{W}{P})  $ \\
\end{spacing}

\begin{spacing}{0.2}
\STATE {\textsc{\textbf{unfold $F_{B}$:}}}\hspace{0.1cm} ${F_{B}}':\textcolor{deepgreen}{(B,L,D)} \gets \textbf{unfold}(F_{B})$ 
\end{spacing}

\begin{spacing}{0.2}
\STATE {\textsc{\textbf{surround traverse:}}}\hspace{0.1cm} $F_{c}:\textcolor{deepgreen}{(B,L,D)} \gets$ \textbf{clock wise traverse} (${F_{B}}'$) \\
$F_{cc}:\textcolor{deepgreen}{(B,L,D)} \gets$ \textbf{counter clock wise traverse} (${F_{B}}'$) 
\end{spacing}

\begin{spacing}{0.4}
\STATE \textbf{Three scanning methods:}  $S_{f} \gets forward$, $S_{fs} \gets  forward\: surround$, $S_{bs} \gets  backward\: surround $
\end{spacing}

\begin{spacing}{0.2}
\STATE $x:\textcolor{deepgreen}{(B,L,d)} \gets \textbf{Linear}^{x}(\textbf{Norm}(F_{B})) $  \\
$z:\textcolor{deepgreen}{(B,L,d)} \gets \textbf{Linear}^{z}(\textbf{Norm}(F_{B})) $
\end{spacing}

\begin{spacing}{0.2}
\STATE $/* process\:  with\:  different\:  scan\:  direction */$
\end{spacing}

\begin{spacing}{0.4}
\STATE \textbf{for} $s \in \left \{  S_{f}, S_{fs}, S_{bs}\right \} $
\textbf{do}
\end{spacing}

\begin{spacing}{0.2}
\STATE \hspace{0.3cm} $x_{s}:\textcolor{deepgreen}{(B,L,d)} \gets \textbf{SiLU}($\textbf{$Conv1d_{s}$}$(x))$   \\
\hspace{0.3cm} $B_{s}:
\textcolor{deepgreen}{(B,L,n)} \gets \textbf{Linear}_{s}^{B}(x) $  \\
\end{spacing}

\begin{spacing}{0.2}
\STATE 
\hspace{0.3cm} $C_{s}:\textcolor{deepgreen}{(B,L,n)} \gets \textbf{Linear}_{s}^{C}(x) $  
\end{spacing}

\begin{spacing}{0.2}
\STATE 
\hspace{0.3cm} $\Delta_{s}:\textcolor{deepgreen}{(B,L,d)} \gets \tau_{\Delta}(\textbf{$Paramer_{s}$}+\textbf{Linear}_{\Delta}(x)) $  
\end{spacing}

\begin{spacing}{0.2}
\STATE 
\hspace{0.3cm} $\bar{A} _{s}:\textcolor{deepgreen}{(B,L,d,n)} \gets \Delta_{s} \otimes $ $\textbf{Paramer}_{s}^{A}$  
\end{spacing}

\begin{spacing}{0.2}
\STATE 
\hspace{0.3cm} $\bar{B} _{s}:\textcolor{deepgreen}{(B,L,d,n)} \gets \Delta_{s} \otimes B_{s} $  
\end{spacing}

\begin{spacing}{0.2}
\STATE 
\hspace{0.3cm} ${y}_{s}:\textcolor{deepgreen}{(B,L,d)} \gets \textbf{SSM}(\bar{A}_{s},\bar{B}_{s},C_{s})(x_{s})$  
\end{spacing}

\STATE  \textbf{end for}

\begin{spacing}{1.2}
\STATE 
\end{spacing}

\begin{spacing}{0.2}
\STATE 
\hspace{0.3cm} ${y}'_{f}:\textcolor{deepgreen}{(B,L,d)} \gets y_{f}\odot \textbf{SiLU}(z)$  
\end{spacing}

\begin{spacing}{0.4}
\STATE 
\hspace{0.3cm} ${y}'_{fs}:\textcolor{deepgreen}{(B,L,d)} \gets y_{fs}\odot \textbf{SiLU}(z)$  
\end{spacing}

\begin{spacing}{0.4}
\STATE 
\hspace{0.3cm} ${y}'_{bs}:\textcolor{deepgreen}{(B,L,d)} \gets y_{bs}\odot \textbf{SiLU}(z)$  
\end{spacing}

\begin{spacing}{0.2}
\STATE \hspace{0.3cm}  $/*residual\: skip\: connection\: */$
\end{spacing}

\begin{spacing}{0.4}
\STATE 
\hspace{0.3cm} $F_{EB}:\textcolor{deepgreen}{(B,L,D)} \gets$ $\textbf{Linear}^{EB}$$({y}'_{f} + {y}'_{fs} + {y}'_{bs}) + F_{B} $  
\end{spacing}

\end{algorithmic}
\label{alg1}
\end{algorithm}

\noindent vehicle centerness. The entire process can be represented as follows:

\begin{equation}
\label{eq_11}
C_{mask}=\sigma(F_{max}(C_{F})\textcircled{c} F_{mean}(C_{F}))
\end{equation}

Where $F_{max}$ and $F_{min}$ represent max pooling and average pooling, respectively. \textcircled{c} denotes concatenation along the channel dimension, and $\sigma$ is the Sigmoid operation.

In the PV space, we introduce Multi-view deformable cross attention based on the deformable attention in the BEVFormer \cite{li2022bevformer}. Whereas in our framework, we replace the predefined queries with confidence heatmap output by the center head $Q_{c} \in \mathbb{R}^{(B, Z\ast X, C)}$, Where $B$ represents batch size, $Z$ and $X$ are left-right and front-back axes of BEV space, respectively, and $C$ denotes the number of channels in the confidence heatmap. The multi-view features $M_{F}=\left \{ F_{i}  \right \}_{i=1}^{n}$ are used as the key $K$ and value $V$, where $F_{i}$ is the PV feature from the $ith$ view. $n$ is the number of all camera views. $\mathcal{P}$ for positional embeddings. Multi-view image features can be represented as $\delta_{1},\delta_{2},\dots \delta_{n}$, which we combine into a single vector $\delta=\left [ \delta_{1},\delta_{2},\dots \delta_{n} \right ]$. For center queries, $Q_{c}\supset [\phi_{1},\phi_{2},\dots, \phi_{n} ]$. The specific formula is expressed as follows:

\begin{equation}
\label{eq_12}
{Q}'_{c}=w_{q} \cdot \left \{[\phi_{1},\mathcal{P}_{1}], [\phi_{2},\mathcal{P}_{2}],\dots,[\phi_{n},\mathcal{P}_{n}]\right \} 
\end{equation}

\begin{equation}
\label{eq_13}
K=w_{k} \cdot [\delta_{1},\delta_{2},\dots \delta_{n}]
\end{equation}

\begin{equation}
\label{eq_13}
V=w_{v} \cdot [\delta_{1},\delta_{2},\dots \delta_{n}]
\end{equation}

Through multi-view deformable cross attention $MCA$, $Q_{c}$ inquires about the vehicle centerness information from the multi-view image features $M_{F}$, which enables the model to focus more on the spatial regions where vehicles are located in the image, thereby better distinguishing between foreground and background. The specific formula can be expressed as follows:

\begin{equation}
\label{eq_13}
MCA({Q}'_{c},M_{F})=softmax\left ( \frac{{Q}'_{c}K^{T} }{\sqrt{d_{k}} } \right ) 
\end{equation}

\subsection{ Loss Function}
Our OE-BevSeg includes a multi-task prediction head. In addition to the segmentation head, we introduce auxiliary task heads to predict vehicle instance centerness and offset. The centerness indicates the probability of finding the vehicle center \cite{wang2021fcos3d}, which is represented by the 2D Gaussian distribution and ranges from zero to one. The offset head output is a vector field, which points to the center of the instance.

The segmentation head uses the cross-entropy loss function, the center head is supervised using the $L2$ loss, and the offset head uses the $L1$ loss. To balance multi-task learning, we employ learnable weights based on uncertainty \cite{kendall2018multi} to balance the three loss functions. The specific formulae are as follows:

\begin{equation}
\label{eq_13}
L_{seg} = -\frac{1}{N} \sum_{i=1}^{N} y_{i,gt} log(P_{i,pred})
\end{equation}

\begin{equation}
\label{eq_13}
L_{cen} = \frac{1}{N} \sum_{i=1}^{N}(c_{i,gt} -c_{i,pred})^{2} 
\end{equation}

\begin{equation}
\label{eq_13}
L_{offt} = \frac{1}{N} \sum_{i=1}^{N}  \left \|  offt_{i,gt} - offt_{i,pred} \right \|_{1} 
\end{equation}

\begin{equation}
\label{eq_13}
L_{total} = L_{seg}\ast w_{CE} + L_{cen}\ast w_{L2} + L_{offt}\ast w_{L1}
\end{equation}

Where $N$ represents the total number of pixels, and 
$P$ is the probability of predicting the $ith$ pixel as a vehicle. $w_{CE}$, $w_{L2}$ and $w_{L1}$ are the learnable uncertainty weights corresponding to the respective loss functions.

\section{EXPERIMENT}
\subsection{\bf Dataset}
In this study, we conducted extensive experiments on the challenging nuScenes \cite{nuscenes2019} urban scenes dataset to demonstrate the effectiveness of our OE-BEVSeg in large-scale autonomous driving scenarios. All data were collected using 6 cameras, 1 LiDAR, and 5 radar sensors, and then sampled at a well-synchronized frequency of 2Hz for the keyframes. The data were collected from 1,000 complex driving scenarios in Boston and Singapore, with 850 scenarios used for training and 150 for testing. There are 28,130 samples in the training set, while the validation set comprises 6,019 samples. We conducted research on BEV segmentation specifically for vehicle objects and performed experiments combining radar/LiDAR multimodal data. The "vehicle" superclass consists of eight categories, i.e., bicycle, bus, car, construction vehicle, emergency vehicle, motorcycle, trailer, and truck. We use the intersection-over-union (IOU) as  evaluation metric.

\begin{table*}[]
\begin{center}
\renewcommand{\arraystretch}{1.3 }
\caption{Comparative experiments with state-of-the-art BEV vehicle segmentation methods.}
\begin{tabular}{ccccccccc}
\toprule \toprule
\textbf{Method }    & \textbf{Public Year}   &\textbf{Backb.}    & \textbf{Lifting}       & \multicolumn{1}{c}{\textbf{Temp.}}    & \textbf{Modalities} & \textbf{Batchsize} & \textbf{Resolution} & \textbf{IOU}  \\
\midrule
FISHING\cite{hendy2020fishing}      & CVPR'2020     &EN-b4     & MLP           &                              & RGB    & -         & 192$\times$320    & 30.0 \\
Lift, Splat\cite{philion2020lift}    & ECCV'2020     &EN-b0     & Depth splat   &                              & RGB    & 4         & 128$\times$352  & 32.1 \\
\multirow{2}{*}{FIERY\cite{hu2021fiery}} & \multirow{2}{*}{CVPR'2021} & \multirow{2}{*}{EN-b4} & \multirow{2}{*}{Depth splat} &  & RGB & 12  & 224$\times$480  & 35.8 \\
    &            &                    &                  & $\checkmark$                 & RGB+time     & 12        & 224$\times$480  & 38.2 \\
CVT\cite{zhou2022cross}          & CVPR'2022     &EN-b4  & Attention       &               & RGB          & 16        & 224$\times$448  & 36.0   \\
\multirow{2}{*}{TIIM\cite{saha2022translating}}  & \multirow{2}{*}{ICRA'2022} & \multirow{2}{*}{RN-50}  & \multirow{2}{*}{Ray attn.}  &   & RGB & 8  & 900 $\times$1600  & 38.9 \\
                       &                   &                             & & $\checkmark$  & RGB+time     & 8         & 900$\times$1600 & 41.3   \\

\multirow{2}{*}{BEVFormer\cite{li2022bevformer}}  & \multirow{2}{*}{ECCV'2022}     & \multirow{2}{*}{RN-101}  & \multirow{2}{*}{Def. Attn.}   &      & RGB          & 1         & 900$\times$1600  & 44.4   \\
&                           &             &                                & $\checkmark$  & RGB+time     & 1    & 900$\times$1600  & 46.7  \\
FedBEVT\cite{song2023fedbevt}     & T-IV-2023     &Trans.        & Attention                     & $\checkmark$  & RGB          & 4   & - & 35.4  \\ 

BAEFormer\cite{pan2023baeformer}   & CVPR'2023     &EN-b4        & Attention                      &               & RGB          & 16   & 224$\times$480 & 41.0  \\ 
PETRV2\cite{liu2023petrv2}     & ICCV'2023     &V2-99        & Attention                & $\checkmark$         & RGB         & 1   & 900$\times$1600 & 46.3 \\  
SimpleBEV\cite{harley2023simple}   & ICRA'2023     &RN-101             & Bilinear                 &               & RGB          & 40   & 448$\times$800 & 47.4 \\                 
UIF-BEV\cite{ren2024uif}     & T-IV-2024     &Trans.        & Attention               & $\checkmark$  & RGB+time      & 4   & 224$\times$448  & 36.0  \\ 
\multirow{3}{*}{PointBEV\cite{chambon2024pointbev}}     & \multirow{3}{*}{CVPR'2024}    & RN-50    & \multirow{3}{*}{Bilinear}   &   & RGB   &28 & 448 × 800 & 47.0       \\
 &                             & EN-b4    &    &       & RGB & 28 & 448$\times$800 & 47.6                     \\
 &                              & EN-b4   &    & $\checkmark$    & RGB+time   &28  & 448$\times$800   & 48.7  \\
\midrule
Ours              & OURS         & V2Res-Net       & Bilinear    &    & RGB    & 8$\times$5  & 448$\times$800     & \textbf{52.6} \\   
Ours              & OURS         & V2Res-Net       & Bilinear    &  & RGB+Radar  & 8$\times$5   & 448$\times$800  & \textbf{58.0}\\   
Ours              & OURS         & V2Res-Net       & Bilinear    &  & RGB+LiDAR  & 8$\times$5   & 448$\times$800  & \textbf{65.3} \\   
\bottomrule
\bottomrule
\end{tabular}
\end{center}
\end{table*}

\subsection{\bf Implementation Details}
For the experimental implementation, We used the PyTorch framework and trained on four 40 GB NVIDIA A100 GPUs for a total of 20,000 iterations. The height and width of the input surround-view images are 448 and 800, respectively. The learning rate was set to 5e-4, using the Adam-W optimizer and the one-cycle learning rate schedule \cite{smith2019super}. The resolution of the BEV space is $200 \times 200$, with a channel dimension of 128, corresponding to a perception range of $100m \times 100m$. The encoder of our OE-BevSeg network adopts the first three stages of VOVNetV2, and the decoder uses the first three layers of ResNet18 as the backbone. Following the report of \cite{harley2023simple}, a larger batch size yields superior results. With limited GPU memory, we accumulate the forward and backward over 5 iterations before performing a parameter update during training. This approach leads to more stable gradient updates and better model performance. Without increasing memory usage, we achieved training with a batch size of 40.

\subsection{\bf State-of-the-art comparison}

We compared our approach with other competitive algorithms for BEV segmentation, including recent most advanced methods such as PETRv2\cite{liu2023petrv2}, SimpleBEV \cite{harley2023simple}, and PointBEV \cite{chambon2024pointbev}. To ensure fair comparison, we also listed the viewpoint lift methods, whether temporal information was used, and the resolution used during training for each algorithm. As shown in Table \uppercase\expandafter{\romannumeral+1}, our OE-BEVSeg demonstrates superior performance. Compared to SimpleBEV, which uses the same perspective transformation method and resolution, our method outperforms by $5.2\%$ in IOU under camera-only conditions. Additionally, compared to the latest PointBEV, which utilizes temporal information, we achieve a $3.9\%$ improvement in IOU. Compared to TIIM, BEVFormer and PETRv2, which use the original image resolution of $900 \times 1600$, our method significantly outperforms these high-resolution training and testing algorithms even when the input is at $1/4$ resolution. Specifically, our method surpasses PETRv2 by $6.3\%$ in IOU. This substantial improvement in BEV segmentation performance demonstrates the effectiveness of our approach. Additionally, Table \uppercase\expandafter{\romannumeral+1} illustrates the superiority of our multi-modal branches. By fusing radar data, our method further improves IOU by $5.4\%$ compared to the camera-only approach. When integrating LiDAR data, the IOU reaches $65.3\%$, exceeding the camera-only approach by a large margin about $12.7\%$.

\begin{table*}[]
\begin{center}
\renewcommand{\arraystretch}{1.5}
\caption{Ablation studies on our proposed Environment-aware BEV Compressor and Center-Informed Object Enhancement block.}
\begin{tabular}{cccc|ccc}
\toprule \toprule
\multicolumn{4}{c}{\textbf{Method}}  & \multicolumn{3}{c}{\textbf{Metric}}   \\ \midrule
\multicolumn{2}{c}{EBC Block} & \multicolumn{2}{c}{CIOE Block} &        & \multicolumn{1}{c}{Without Flitering} & \multicolumn{1}{c}{Visibility Filtering} \\ \midrule
Normal Scan    & Surround Scan    & BEV Enhance    & PV Enhance    & Params(M)$\downarrow$ & IOU$\uparrow$           & IOU$\uparrow$                         \\ \midrule 
               &                  &                &               &  62.27             &  43.68                  & 50.53             \\
$\checkmark$   &                  &                &               &  68.94             &  44.26                  & 51.33              \\
$\checkmark$   &                  & $\checkmark$   &               &  69.23             &  44.34                  & 51.64            \\
$\checkmark$   & $\checkmark$     &                &               &  69.69             &  44.40                  & 51.75            \\
$\checkmark$   &  $\checkmark$    & $\checkmark$   &               &  69.98             &  44.37                  & 52.10            \\
$\checkmark$   & $\checkmark$     & $\checkmark$   & $\checkmark$  & \textbf{80.37}     & \textbf{ 44.98}         & \textbf{52.65}    \\
\bottomrule \bottomrule  
\end{tabular}
\end{center}
\end{table*}


\subsection{\bf Ablation Studies}
In this section, we conduct extensive ablation studies on the core ingredients proposed in this paper: the EBC Block and CIOE Block, the combination of loss functions, different image resolutions, the combination of different modal data, and different vehicle perception distances.

\begin{table}[]
\begin{center}
\renewcommand{\arraystretch}{1.3}
\caption{Ablation experiments on different loss functions.}
\begin{tabular}{cccc}
\toprule 
\multicolumn{3}{c}{\textbf{Vehicle seg. IoU (↑)}} & \textbf{Metric} \\
\midrule
CE loss   & Center loss   & Offset loss  & IOU(↑)    \\
\midrule
$\checkmark$           &  $\checkmark$        &                            &  47.42     \\
$\checkmark$           &                      &  $\checkmark$              &  46.93      \\
$\checkmark$           &  $\checkmark$        &  $\checkmark$              &  \textbf{48.04}       \\
\bottomrule      
\end{tabular}
\end{center}
\end{table}

\begin{table}[]
\begin{center}
\renewcommand{\arraystretch}{1}
\caption{Ablation experiments on different image resolutions.}
\begin{tabular}{cccc}
\toprule 
\textbf{Vehicle seg. IoU (↑) }                    &                               & \multicolumn{2}{c}{\textbf{Input Shape}}                 \\
\midrule
Method                                    & Backb.                   & 224$\times$480       & 448$\times$800                \\
\midrule
FIERY\cite{hu2021fiery}                                     & EN-b4                         & 39.8                 & -                        \\
CVT\cite{zhou2022cross}                                       & EN-b4                         & 36.0                 & 37.7                     \\
LaRa \cite{bartoccioni2023lara}                                     & EN-b4                         & 38.9                 & -                        \\
BEVFormer \cite{li2022bevformer}                                 & RN-50                         & 42.0                 & 45.5                     \\
BAEFormer \cite{pan2023baeformer}                                & EN-b4                         & 38.9                 & 41.0                     \\
SimpleBEV  \cite{harley2023simple}\                              & RN-50                         & 43.0                 & 46.6                     \\
\multirow{2}{*}{PointBEV\cite{chambon2024pointbev}}                 & RN-50                         & 44.1                 & 47.7                     \\
                                          & EN-b4                         & 44.7                 & 48.7                     \\
\midrule
\makecell{OE-BEVSeg \\ (Ours)}       & V2Res-Net     &  \textbf{ 47.6}     &  \textbf{52.6}        \\  
\bottomrule
\end{tabular}
\end{center}
\end{table}

\begin{table}[]
\begin{center}
\renewcommand{\arraystretch}{1.1}
\caption{Comparative experiments on different modal data.}
\begin{tabularx}{\linewidth}{cccccc}
\toprule 
\textbf{Vehicle. seg. IoU (↑)} & \multicolumn{4}{c}{\textbf{Modalities}}                       &   \\ \hline
Method                         & Radar & Lidar  & RGB  & IOU(↑)  \\
\midrule
\multirow{2}{*}{FISHING\cite{hendy2020fishing}}      &                 &                & $\checkmark$    & 30.0 \\
                              &                 & $\checkmark$   & $\checkmark$   & $44.3_{\uparrow14.3}$  \\
\midrule
\multirow{2}{*}{Lift, Splat\cite{philion2020lift}}  &                 &                & $\checkmark$    & 32.1 \\
                              &                 & $\checkmark$   & $\checkmark$   & $44.5_{\uparrow12.4}$ \\
\midrule
\multirow{3}{*}{SimpleBEV\cite{harley2023simple}}    &                 &                & $\checkmark$     & 47.4 \\
                              & $\checkmark$    &                & $\checkmark$    & $55.7_{\uparrow8.30}$ \\
                              &                 & $\checkmark$   & $\checkmark$    & $60.8_{\uparrow13.4}$\\
\midrule
\multirow{3}{*}{\makecell{OE-BEVSeg \\ (Ours)}}   &        &     & $\checkmark$     & 52.6     \\
                              & $\checkmark$    &                & $\checkmark$     & $\mathbf{58.0_{\uparrow5.4}}$     \\
                              &                 & $\checkmark$   & $\checkmark$     & $\mathbf{65.3_{\uparrow12.7}}$      \\ 
\bottomrule

\end{tabularx}
\end{center}
\end{table}

\subsubsection{\bf Effectiveness of EBC block}
The Environment-aware BEV Compressor is our proposed global long-distance environment awareness module based on Mamba. Specifically, we designed a surround scan mechanism. As shown in Table \uppercase\expandafter{\romannumeral+2}, we conducted ablation experiments on the normal scan and surround scan methods. The visibility filtering means that if a car is not visible from any camera viewpoint, we discard it. Without filtering means all annotated vehicles are considered. It can be observed that with the use of the visibility filter, the introduction of the EBC module increases the IOU by $0.8\%$ compared to the baseline. Without filtering, the increase is $0.58\%$. When we use both normal scan and surround scan and use visibility filter, the IOU improvement is $1.22\%$.

\begin{table*}[]
\begin{center}
\renewcommand{\arraystretch}{1.5}
\caption{Ablation study on the accuracy of vehicle perception with increasing distance from the ego-vehicle.}
\setlength{\tabcolsep}{2.5mm}{ 
\begin{tabular}{cccccc|cccccccc}
\toprule \toprule
\multirow{3}{*}{\textbf{Method}} & \multirow{3}{*}{\textbf{Modalities}} & \multicolumn{4}{c}{\textbf{Visibility Filtering}} & \multicolumn{4}{c}{\textbf{Without Flitering}}\\ \cline{3-10} 
                                 &               & \multicolumn{4}{c}{Vehicle Perception Range}  & \multicolumn{4}{c}{Vehicle Perception Range} \\
                                 &               & 0-20m  & 20-35m  & 35-50m & 0-50m  & 0-20m  & 20-35m  & 35-50m & 0-50m\\
\midrule
\multirow{2}{*}{SimpleBEV}       & C                     &  65.7       &  44.9        &  27.3        &   47.4   &  62.0 & 38.6 & 22.5  &  40.8  \\
                                 & C+R                   &  70.1       &  53.6        &  39.1        &   55.5   &  66.1 & 45.4 & 31.4  &  47.6   \\
\midrule                      
\multirow{3}{*}{\makecell{OE-BEVSeg \\ (Ours)}} & C      &  70.6       &  51.0        &  32.9       & 52.6   & 66.2  & 42.9 & 26.8 & 44.9  \\
                                 & C+R                   &  72.8       &  56.8        &  40.6       & 58.0   & 68.4  & 47.1 & 32.5 & 49.1 \\
                                 & C+L                   &  \textbf{78.3}       &  \textbf{64.9}        &  \textbf{48.7}       & \textbf{65.3}   & \textbf{73.7}  & \textbf{54.2} & \textbf{38.5} & \textbf{55.4}  \\
\bottomrule \bottomrule
\end{tabular}}
\end{center}
\end{table*}

\begin{figure*}
\centering
\includegraphics[width=7in]{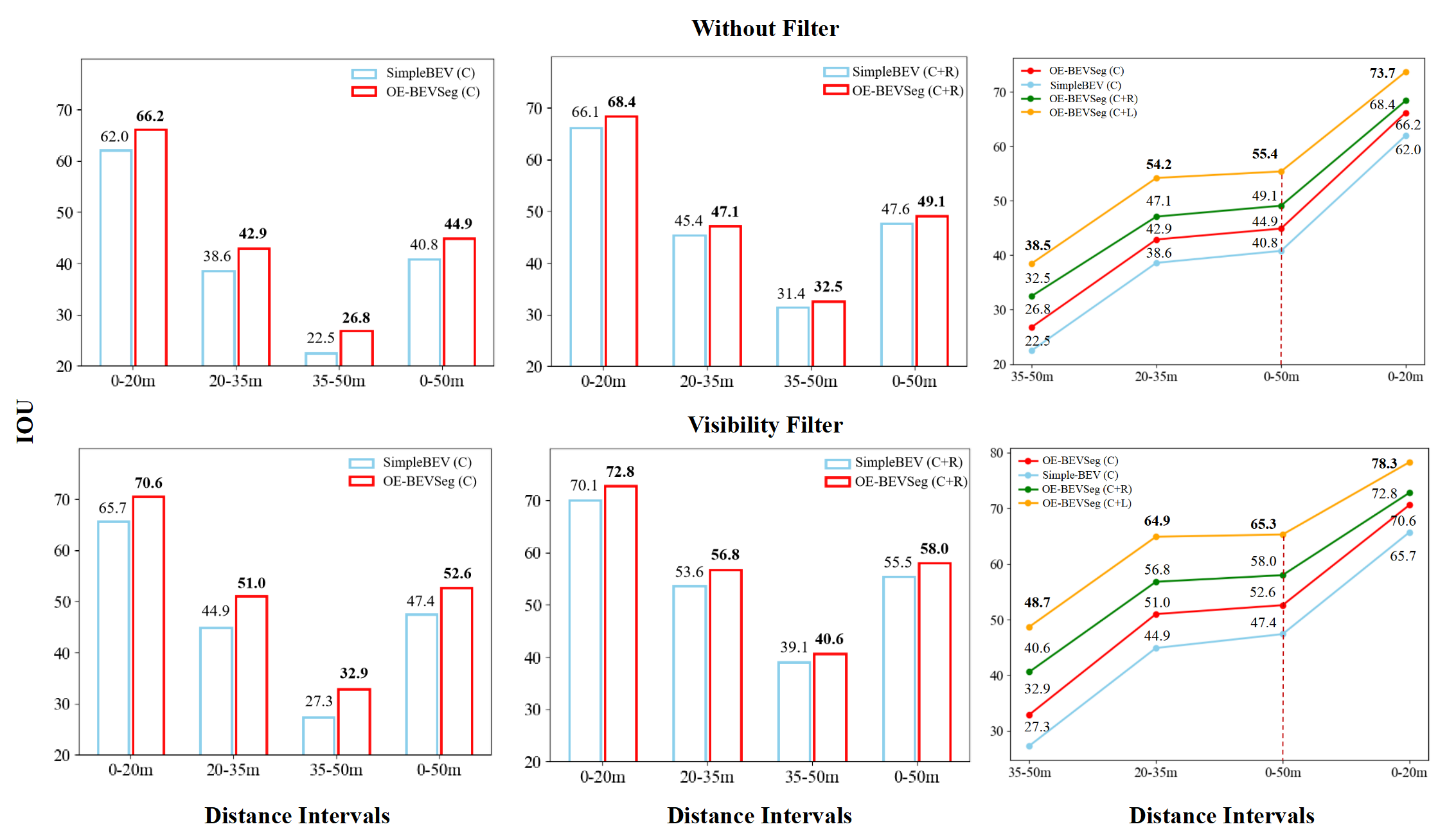}
\caption{Visualization demonstrating the accuracy of vehicle perception with increasing distance from the ego-vehicle. We conducted comparative experiments with SimpleBEV using different data fusions (RGB, Radar).}
\label{fig_6}
\end{figure*}

\begin{figure*}
\centering
\includegraphics[width=7in]{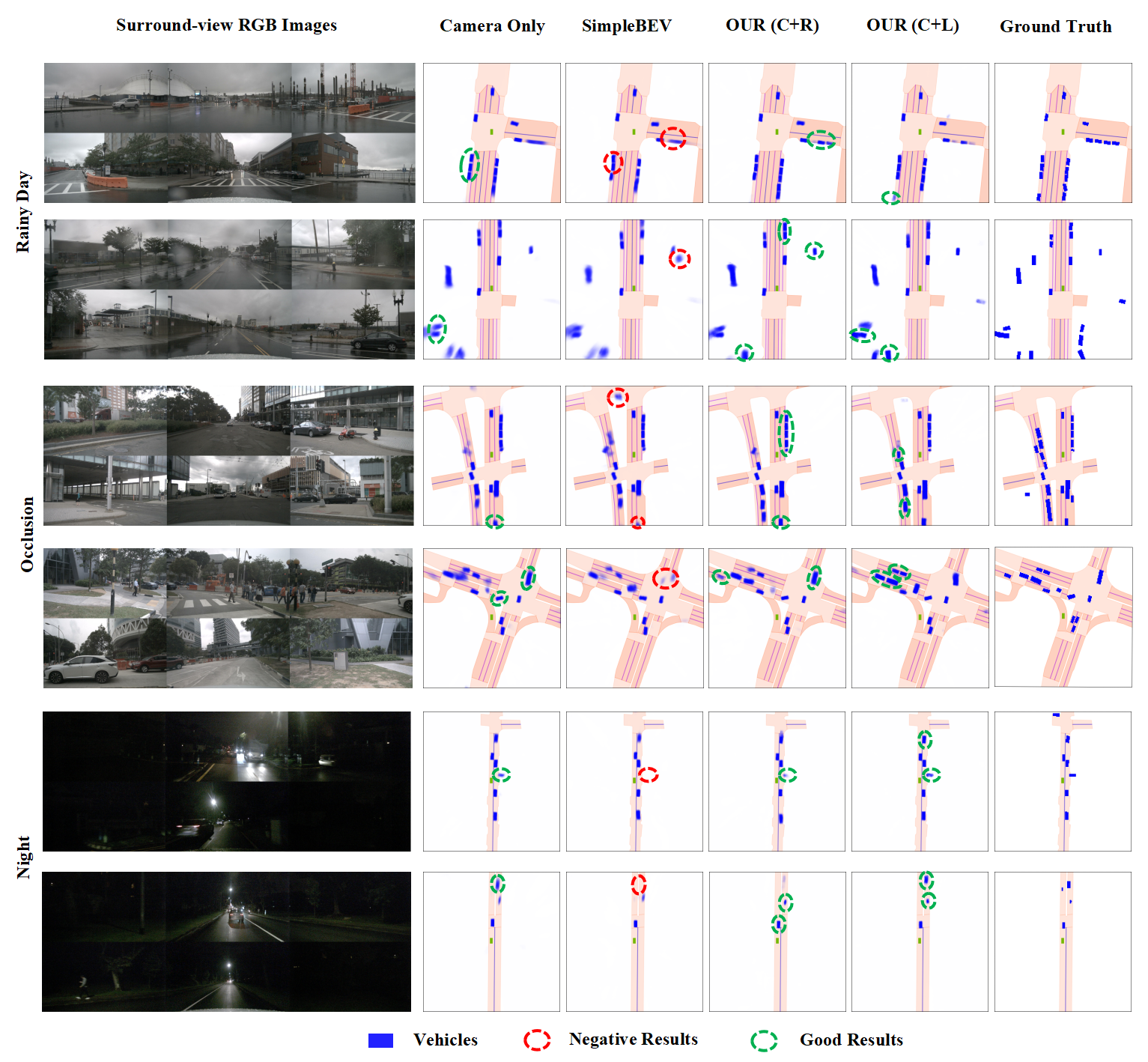}
\caption{Visualization of BEV segmentation results under different weather conditions, times of day, and occlusion challenge scenarios. We present a comparison of the segmentation results of our proposed OE-BevSeg in different modalities as well as SimpleBEV.}
\label{fig_7}
\end{figure*}

\begin{figure*}
\centering
\includegraphics[width=5in]{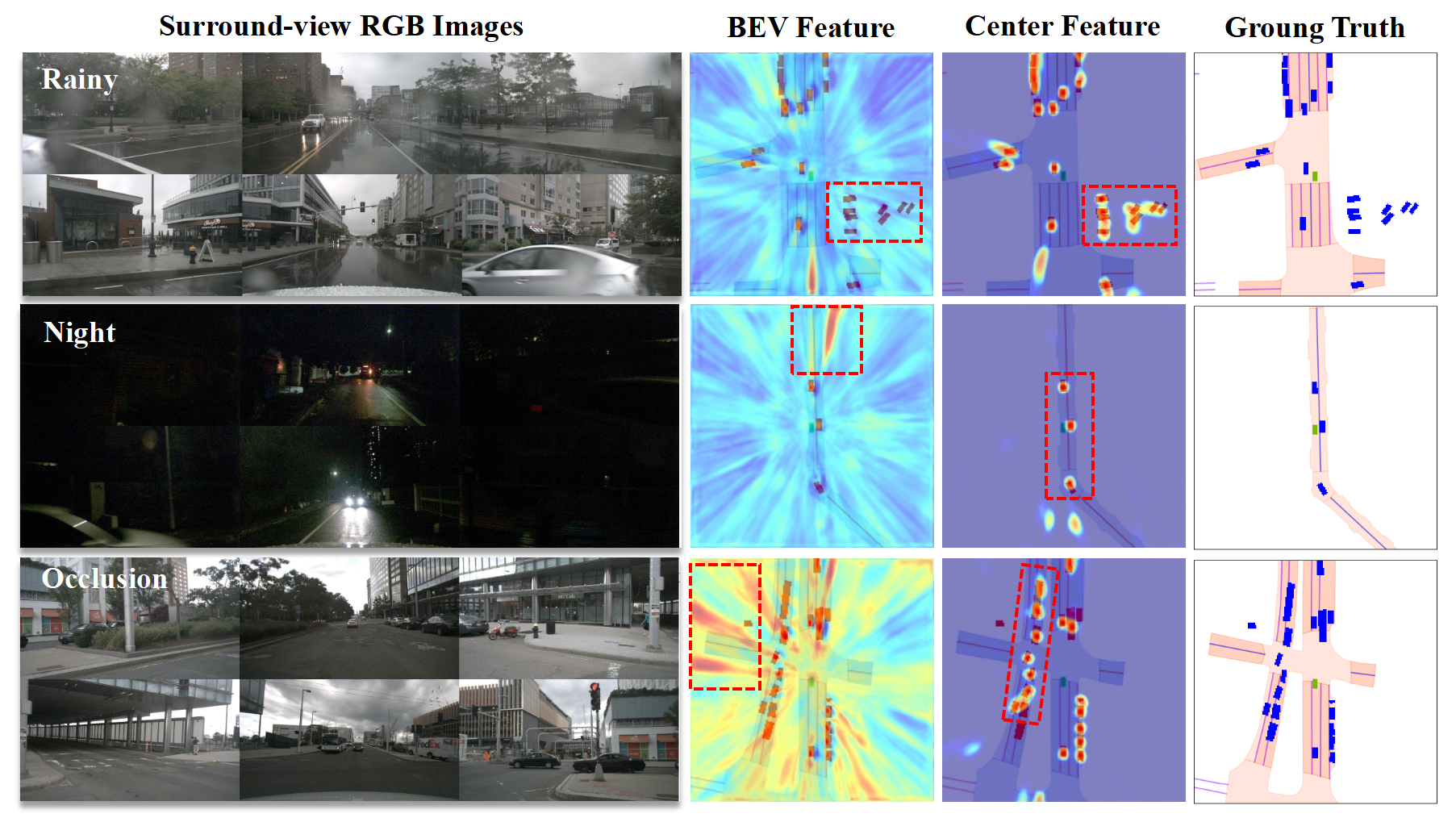}
\caption{The left image shows the surround-view RGB images in different scenes, while the right side displays the heatmap visualizations of the BEV feature and center feature. The far-right image shows the ground truth of vehicle mask, with the blue areas indicating vehicles. The BEV feature is obtained from the output of the last layer of the decoder, and the center feature is from the center head branch.}
\label{fig_8}
\end{figure*}

\begin{figure*}
\centering
\includegraphics[width=6in]{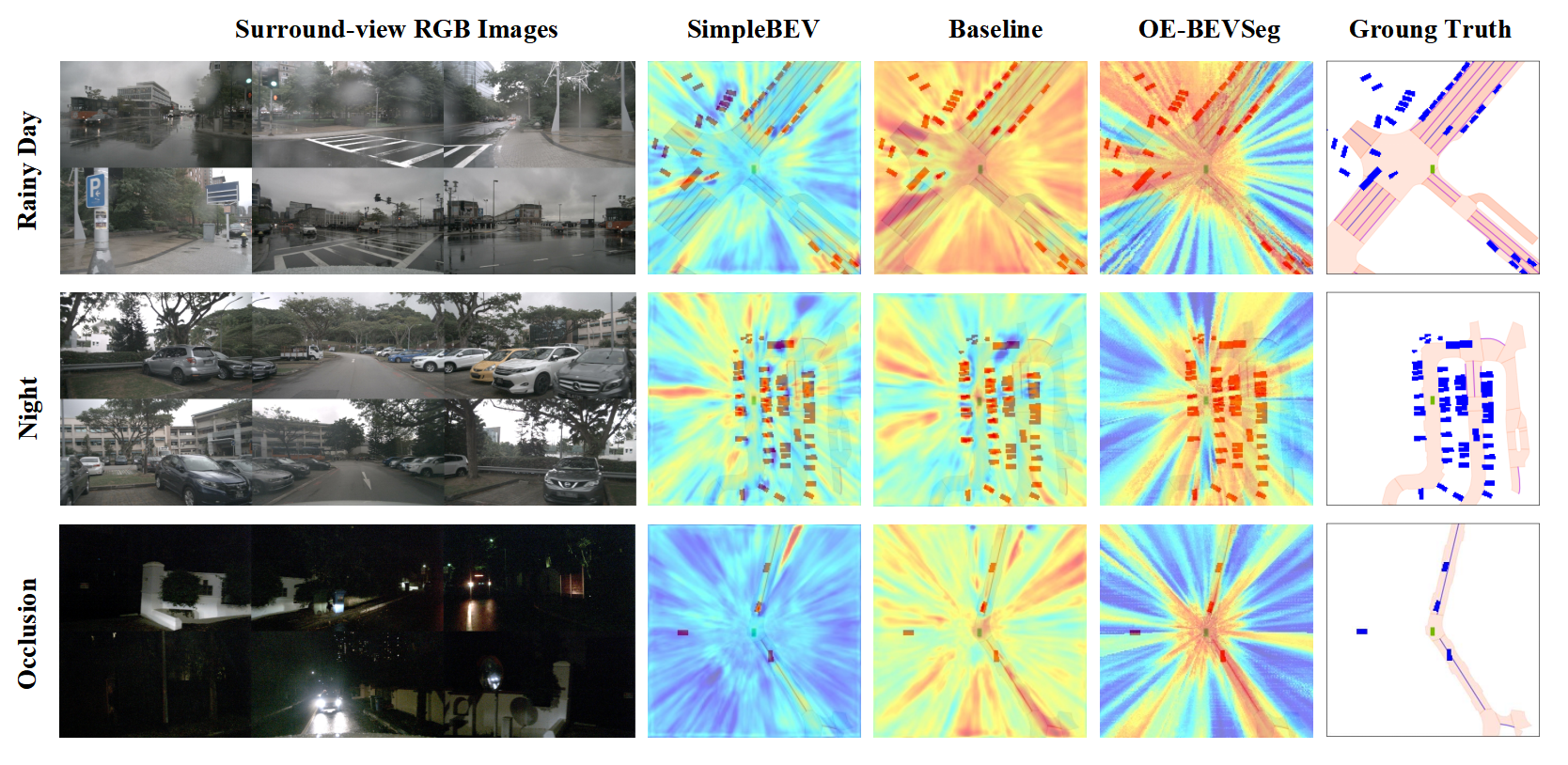}
\caption{The left image shows the surround-view RGB images in different scenes. The right side shows the heatmap visualizations of the features from SimpleBEV, V2Res-Net baseline, and OE-BevSeg, as well as the ground truth of the vehicle mask. The feature maps are obtained from the output of the last layer of the decoder.}
\label{fig_9}
\end{figure*}

\subsubsection{\bf Effectiveness of CIOE block}

The center-informed object enhancement is proposed to better explicitly utilize vehicle instance centerness, enhancing and enriching the detail information of the target object in the segmentation results, and effectively distinguishing between the foreground and background. As shown in Table \uppercase\expandafter{\romannumeral+2}, we achieve different levels of improvement by enhancing the BEV space and PV space using the CIOE block. Additionally, the experimental results demonstrate that using both the CIOE block and the EBC block yields the greatest benefits. After adding the CIOE block to the network, the model's IOU improved by $0.9\%$. Using both the CIOE block and the EBC block together, compared to the backbone, resultes in a $2.12\%$ IOU improvement.

\subsubsection{\bf Effectiveness of loss functions}

In this paper, we propose a multi-task prediction head corresponding to three different losses: the cross-entropy loss function, center loss, and offset loss. As shown in Table \uppercase\expandafter{\romannumeral+3}, at the beginning of our experiments, we performed ablation studies on the loss functions based on the baseline. It can be seen that introducing center loss on top of the cross-entropy loss function achieved better results than the offset loss. When we used all three losses simultaneously, the IOU reached a maximum of $48.04\%$. This also demonstrates the effectiveness of our approach to enhancing target object information using vehicle centerness.

\subsubsection{\bf Ablation experiments with image resolution}

The input image resolution has a significant impact on segmentation tasks, with larger resolutions generally yielding better segmentation results. In Table \uppercase\expandafter{\romannumeral+4}, we compared advanced models using two image resolutions: $224 \times 480$ and $448 \times 800$. These models include CVT, BEVFormer, SimpleBEV, and PointBeV. The experimental results demonstrate that our OE-BevSeg achieved the best IOU regardless of whether a larger or smaller image resolution was used. Furthermore, OE-BevSeg outperformed other algorithms by a greater margin at higher resolutions. When the resolution was $448 \times 800$, the IOU exceeded that of SimpleBEV by $6\%$ and PointBEV by $3.9\%$. These results confirm the importance of high-resolution images for BEV segmentation tasks and also demonstrate the robustness of our method to different resolution inputs.

\subsubsection{\bf Effectiveness of multimodal fusion}

In this paper, we employed a multi-modal fusion branch to integrate multi-view RGB images with radar/LiDAR data, significantly enhancing the performance of BEV segmentation. As shown in Table \uppercase\expandafter{\romannumeral+5}, we conducted comparative experiments with advanced BEV semantic segmentation algorithms that also use multi-modal data. The inclusion of additional auxiliary modal data resulted in substantial performance improvements across these algorithms. Compared to the best-performing SimpleBEV, our method achieved the highest IOU of $58.0\%$ when fusing radar data, surpassing SimpleBEV by $2.3\%$. When fusing LiDAR data, our method also achieved the highest IOU of $65.3\%$, surpassing SimpleBEV by $4.2\%$. The fusion of LiDAR data demonstrated better potential performance compared to radar data. Our OE-BevSeg showing a $5.4\%$ improvement when using radar data in the fusion branch, and a substantial $12.7\%$ improvement when using LiDAR data.

\subsubsection{\bf Ablation experiments with vehicle perception distance}

As shown in Table \uppercase\expandafter{\romannumeral+6}, we evaluated the model's performance as the distance from the ego vehicle increases. We divided vehicle perception into three ranges: $\operatorname{0-20m}$, $\operatorname{20-35m}$, and $\operatorname{35-50m}$, and also assessed the overall segmentation performance in $\operatorname{0-50m}$. We conducted relevant ablation experiments under both visibility filtering and without filtering conditions. As depicted in Fig.\ref{fig_6}, it is clear that as the distance intervals increase, the model's performance gradually declines. This also validates the feasibility of using the hierarchical environmental composition in the BEV space as prior knowledge. Our model demonstrated superior performance, surpassing SimpleBEV in every distance interval. Notably, in the challenging long-distance segmentation ($\operatorname{35-50m}$), our OE-BevSeg outperformed SimpleBEV by $4.3\%$ under the without filtering condition and by $5.6\%$ under the visibility filtering condition.

\subsection{\bf Qualitative Results}
As shown in Fig.\ref{fig_7}, we visualized the BEV vehicle segmentation results and selected some challenging scenarios, such as rainy, occlusion, and night. Our OE-BevSeg achieved segmentation results closest to the ground truth, even in these difficult scenarios. The blue mask represents vehicles, the red annotations indicate negative results, and the green annotations indicate good results. It can be seen that OE-BevSeg provides better segmentation details. Compared to SimpleBEV, we reduce the number of true negative samples in night and occlusion scenarios, as well as the number of false negative samples in rainy scenarios. The fusion of multi-modal information, such as radar and LiDAR, enriches the segmentation details, better delineating vehicle contours. Multi-modal information provides valuable cues for measuring scene structure, enhancing the BEV segmentation performance.

To more intuitively demonstrate the benefits of centerness in enhancing object detail information, as shown in Fig.\ref{fig_8}, we visualized the BEV feature and center feature using heatmaps based on the V2Res-Net baseline. We can observe a clear distinction between the visualization of the BEV and center feature. The BEV feature focuses more on global perception but lacks sufficient attention to the target object. In contrast, the center feature focuses more on the vehicle area, showing higher activation values. Therefore, utilizing centerness to supervise and guide the segmentation head enables the model to better distinguish between the foreground and background.

Fig.\ref{fig_9} shows the heatmap visualizations of the BEV features from the decoder of different models. It can be seen that in challenging scenarios, our OE-BevSeg can effectively focus on the vehicle object regions and better understand the environment, demonstrating improved differentiation between relevant areas with target objects and irrelevant areas without targets. This indicates that our model not only achieves better perception and understanding of the BEV long-distance environment but also enriches the information of target objects, thereby achieving superior vehicle segmentation performance.


\section{Conclusion}


In this work, an innovative end-to-end multimodal framework is designed to address two shortcomings in current BEV semantic segmentation methods, particularly in capturing long-distance environmental features and recognizing fine details of target objects. We have proposed two novel techniques, i.e., environment-aware BEV compressor and center-informed object enhancement. By leveraging long-sequence global modeling based on prior knowledge about the BEV surrounding environment, the model’s long-distance environmental perception and understanding are significantly improved. Furthermore, from a local enhancement perspective, the instance centerness information is utilized to enrich detail target object information. Extensive experiments conducted on the nuScenes dataset demonstrate that our approach achieves state-of-the-art results in both camera-only and multimodal fusion BEV vehicle semantic segmentation tasks. In future work, we aim to explore enhancing the computational efficiency of the BEV segmentation algorithm while maintaining high accuracy, enabling better application in autonomous driving scenarios.


\bibliographystyle{IEEEtran}
\bibliography{mybibfile}

\begin{thebibliography}{10}
\providecommand{\url}[1]{#1}
\csname url@samestyle\endcsname
\providecommand{\newblock}{\relax}
\providecommand{\bibinfo}[2]{#2}
\providecommand{\BIBentrySTDinterwordspacing}{\spaceskip=0pt\relax}
\providecommand{\BIBentryALTinterwordstretchfactor}{4}
\providecommand{\BIBentryALTinterwordspacing}{\spaceskip=\fontdimen2\font plus
\BIBentryALTinterwordstretchfactor\fontdimen3\font minus \fontdimen4\font\relax}
\providecommand{\BIBforeignlanguage}[2]{{%
\expandafter\ifx\csname l@#1\endcsname\relax
\typeout{** WARNING: IEEEtran.bst: No hyphenation pattern has been}%
\typeout{** loaded for the language `#1'. Using the pattern for}%
\typeout{** the default language instead.}%
\else
\language=\csname l@#1\endcsname
\fi
#2}}
\providecommand{\BIBdecl}{\relax}
\BIBdecl

\bibitem{teng2023motion}
S.~Teng, X.~Hu, P.~Deng, B.~Li, Y.~Li, Y.~Ai, D.~Yang, L.~Li, Z.~Xuanyuan, F.~Zhu \emph{et~al.}, ``Motion planning for autonomous driving: The state of the art and future perspectives,'' \emph{IEEE Transactions on Intelligent Vehicles}, vol.~8, no.~6, pp. 3692--3711, 2023.

\bibitem{he2022diff}
L.~He, S.~Jiang, X.~Liang, N.~Wang, and S.~Song, ``Diff-net: Image feature difference based high-definition map change detection for autonomous driving,'' in \emph{2022 International Conference on Robotics and Automation (ICRA)}.\hskip 1em plus 0.5em minus 0.4em\relax IEEE, 2022, pp. 2635--2641.

\bibitem{he2024neural}
L.~He, L.~Li, W.~Sun, Z.~Han, Y.~Liu, S.~Zheng, J.~Wang, and K.~Li, ``Neural radiance field in autonomous driving: A survey,'' \emph{arXiv preprint arXiv:2404.13816}, 2024.

\bibitem{han20234d}
Z.~Han, J.~Wang, Z.~Xu, S.~Yang, L.~He, S.~Xu, J.~Wang, and K.~Li, ``4d millimeter-wave radar in autonomous driving: A survey,'' \emph{arXiv preprint arXiv:2306.04242}, 2023.

\bibitem{hu2021fiery}
A.~Hu, Z.~Murez, N.~Mohan, S.~Dudas, J.~Hawke, V.~Badrinarayanan, R.~Cipolla, and A.~Kendall, ``Fiery: Future instance prediction in bird's-eye view from surround monocular cameras,'' in \emph{Proceedings of the IEEE/CVF International Conference on Computer Vision}, 2021, pp. 15\,273--15\,282.

\bibitem{li2022bevformer}
Z.~Li, W.~Wang, H.~Li, E.~Xie, C.~Sima, T.~Lu, Y.~Qiao, and J.~Dai, ``Bevformer: Learning bird’s-eye-view representation from multi-camera images via spatiotemporal transformers,'' in \emph{European conference on computer vision}.\hskip 1em plus 0.5em minus 0.4em\relax Springer, 2022, pp. 1--18.

\bibitem{zhou2022cross}
B.~Zhou and P.~Kr{\"a}henb{\"u}hl, ``Cross-view transformers for real-time map-view semantic segmentation,'' in \emph{Proceedings of the IEEE/CVF conference on computer vision and pattern recognition}, 2022, pp. 13\,760--13\,769.

\bibitem{xu2024surround}
H.~Xu, X.~Zhang, J.~He, Z.~Geng, C.~Pang, and Y.~Yu, ``Surround-view water surface bev segmentation for autonomous surface vehicles: Dataset, baseline and hybrid-bev network,'' \emph{IEEE Transactions on Intelligent Vehicles}, 2024.

\bibitem{sun2022bi}
J.~Sun, J.~Shen, X.~Wang, Z.~Mao, and J.~Ren, ``Bi-unet: A dual stream network for real-time highway surface segmentation,'' \emph{IEEE Transactions on Intelligent Vehicles}, vol.~8, no.~2, pp. 1549--1563, 2022.

\bibitem{harley2023simple}
A.~W. Harley, Z.~Fang, J.~Li, R.~Ambrus, and K.~Fragkiadaki, ``Simple-bev: What really matters for multi-sensor bev perception?'' in \emph{2023 IEEE International Conference on Robotics and Automation (ICRA)}.\hskip 1em plus 0.5em minus 0.4em\relax IEEE, 2023, pp. 2759--2765.

\bibitem{liu2023bird}
J.~Liu, Z.~Cao, J.~Yang, X.~Liu, Y.~Yang, and Z.~Qu, ``Bird's-eye-view semantic segmentation with two-stream compact depth transformation and feature rectification,'' \emph{IEEE Transactions on Intelligent Vehicles}, vol.~8, no.~11, pp. 4546--4558, 2023.

\bibitem{liu2023bevfusion}
Z.~Liu, H.~Tang, A.~Amini, X.~Yang, H.~Mao, D.~L. Rus, and S.~Han, ``Bevfusion: Multi-task multi-sensor fusion with unified bird's-eye view representation,'' in \emph{2023 IEEE international conference on robotics and automation (ICRA)}.\hskip 1em plus 0.5em minus 0.4em\relax IEEE, 2023, pp. 2774--2781.

\bibitem{bertozz1998stereo}
M.~Bertozz, A.~Broggi, and A.~Fascioli, ``Stereo inverse perspective mapping: theory and applications,'' \emph{Image and vision computing}, vol.~16, no.~8, pp. 585--590, 1998.

\bibitem{vaswani2017attention}
A.~Vaswani, N.~Shazeer, N.~Parmar, J.~Uszkoreit, L.~Jones, A.~N. Gomez, {\L}.~Kaiser, and I.~Polosukhin, ``Attention is all you need,'' \emph{Advances in neural information processing systems}, vol.~30, 2017.

\bibitem{zhu2020deformable}
X.~Zhu, W.~Su, L.~Lu, B.~Li, X.~Wang, and J.~Dai, ``Deformable detr: Deformable transformers for end-to-end object detection,'' \emph{arXiv preprint arXiv:2010.04159}, 2020.

\bibitem{huang2021bevdet}
J.~Huang, G.~Huang, Z.~Zhu, Y.~Ye, and D.~Du, ``Bevdet: High-performance multi-camera 3d object detection in bird-eye-view,'' \emph{arXiv preprint arXiv:2112.11790}, 2021.

\bibitem{ren2024uif}
Y.~Ren, L.~Wang, M.~Li, H.~Jiang, C.~Lin, H.~Yu, and Z.~Cui, ``Uif-bev: An underlying information fusion framework for bird's-eye-view semantic segmentation,'' \emph{IEEE Transactions on Intelligent Vehicles}, 2024.

\bibitem{qi2024ocbev}
Z.~Qi, J.~Wang, X.~Wu, and H.~Zhao, ``Ocbev: Object-centric bev transformer for multi-view 3d object detection,'' in \emph{2024 International Conference on 3D Vision (3DV)}.\hskip 1em plus 0.5em minus 0.4em\relax IEEE, 2024, pp. 1188--1197.

\bibitem{gu2023mamba}
A.~Gu and T.~Dao, ``Mamba: Linear-time sequence modeling with selective state spaces,'' \emph{arXiv preprint arXiv:2312.00752}, 2023.

\bibitem{gu2021efficiently}
A.~Gu, K.~Goel, and C.~R{\'e}, ``Efficiently modeling long sequences with structured state spaces,'' \emph{arXiv preprint arXiv:2111.00396}, 2021.

\bibitem{pan2020cross}
B.~Pan, J.~Sun, H.~Y.~T. Leung, A.~Andonian, and B.~Zhou, ``Cross-view semantic segmentation for sensing surroundings,'' \emph{IEEE Robotics and Automation Letters}, vol.~5, no.~3, pp. 4867--4873, 2020.

\bibitem{roddick2020predicting}
T.~Roddick and R.~Cipolla, ``Predicting semantic map representations from images using pyramid occupancy networks,'' in \emph{Proceedings of the IEEE/CVF Conference on Computer Vision and Pattern Recognition}, 2020, pp. 11\,138--11\,147.

\bibitem{peng2023bevsegformer}
L.~Peng, Z.~Chen, Z.~Fu, P.~Liang, and E.~Cheng, ``Bevsegformer: Bird's eye view semantic segmentation from arbitrary camera rigs,'' in \emph{Proceedings of the IEEE/CVF Winter Conference on Applications of Computer Vision}, 2023, pp. 5935--5943.

\bibitem{hendy2020fishing}
N.~Hendy, C.~Sloan, F.~Tian, P.~Duan, N.~Charchut, Y.~Xie, C.~Wang, and J.~Philbin, ``Fishing net: Future inference of semantic heatmaps in grids,'' \emph{arXiv preprint arXiv:2006.09917}, 2020.

\bibitem{kim2023crn}
Y.~Kim, J.~Shin, S.~Kim, I.-J. Lee, J.~W. Choi, and D.~Kum, ``Crn: Camera radar net for accurate, robust, efficient 3d perception,'' in \emph{Proceedings of the IEEE/CVF International Conference on Computer Vision}, 2023, pp. 17\,615--17\,626.

\bibitem{gu2020hippo}
A.~Gu, T.~Dao, S.~Ermon, A.~Rudra, and C.~R{\'e}, ``Hippo: Recurrent memory with optimal polynomial projections,'' \emph{Advances in neural information processing systems}, vol.~33, pp. 1474--1487, 2020.

\bibitem{liu2024vmamba}
Y.~Liu, Y.~Tian, Y.~Zhao, H.~Yu, L.~Xie, Y.~Wang, Q.~Ye, and Y.~Liu, ``Vmamba: Visual state space model,'' \emph{arXiv preprint arXiv:2401.10166}, 2024.

\bibitem{zhu2024vision}
L.~Zhu, B.~Liao, Q.~Zhang, X.~Wang, W.~Liu, and X.~Wang, ``Vision mamba: Efficient visual representation learning with bidirectional state space model,'' \emph{arXiv preprint arXiv:2401.09417}, 2024.

\bibitem{zhao2024rs}
S.~Zhao, H.~Chen, X.~Zhang, P.~Xiao, L.~Bai, and W.~Ouyang, ``Rs-mamba for large remote sensing image dense prediction,'' \emph{arXiv preprint arXiv:2404.02668}, 2024.

\bibitem{ma2024u}
J.~Ma, F.~Li, and B.~Wang, ``U-mamba: Enhancing long-range dependency for biomedical image segmentation,'' \emph{arXiv preprint arXiv:2401.04722}, 2024.

\bibitem{xing2024segmamba}
Z.~Xing, T.~Ye, Y.~Yang, G.~Liu, and L.~Zhu, ``Segmamba: Long-range sequential modeling mamba for 3d medical image segmentation,'' \emph{arXiv preprint arXiv:2401.13560}, 2024.

\bibitem{lee2019energy}
Y.~Lee, J.-w. Hwang, S.~Lee, Y.~Bae, and J.~Park, ``An energy and gpu-computation efficient backbone network for real-time object detection,'' in \emph{Proceedings of the IEEE/CVF conference on computer vision and pattern recognition workshops}, 2019, pp. 0--0.

\bibitem{he2016identity}
K.~He, X.~Zhang, S.~Ren, and J.~Sun, ``Identity mappings in deep residual networks,'' in \emph{Computer Vision--ECCV 2016: 14th European Conference, Amsterdam, The Netherlands, October 11--14, 2016, Proceedings, Part IV 14}.\hskip 1em plus 0.5em minus 0.4em\relax Springer, 2016, pp. 630--645.

\bibitem{hu2018squeeze}
J.~Hu, L.~Shen, and G.~Sun, ``Squeeze-and-excitation networks,'' in \emph{Proceedings of the IEEE conference on computer vision and pattern recognition}, 2018, pp. 7132--7141.

\bibitem{wang2021fcos3d}
T.~Wang, X.~Zhu, J.~Pang, and D.~Lin, ``Fcos3d: Fully convolutional one-stage monocular 3d object detection,'' in \emph{Proceedings of the IEEE/CVF International Conference on Computer Vision}, 2021, pp. 913--922.

\bibitem{kendall2018multi}
A.~Kendall, Y.~Gal, and R.~Cipolla, ``Multi-task learning using uncertainty to weigh losses for scene geometry and semantics,'' in \emph{Proceedings of the IEEE conference on computer vision and pattern recognition}, 2018, pp. 7482--7491.

\bibitem{nuscenes2019}
H.~Caesar, V.~Bankiti, A.~H. Lang, S.~Vora, V.~E. Liong, Q.~Xu, A.~Krishnan, Y.~Pan, G.~Baldan, and O.~Beijbom, ``nuscenes: A multimodal dataset for autonomous driving,'' \emph{arXiv preprint arXiv:1903.11027}, 2019.

\bibitem{philion2020lift}
J.~Philion and S.~Fidler, ``Lift, splat, shoot: Encoding images from arbitrary camera rigs by implicitly unprojecting to 3d,'' in \emph{Computer Vision--ECCV 2020: 16th European Conference, Glasgow, UK, August 23--28, 2020, Proceedings, Part XIV 16}.\hskip 1em plus 0.5em minus 0.4em\relax Springer, 2020, pp. 194--210.

\bibitem{saha2022translating}
A.~Saha, O.~Mendez, C.~Russell, and R.~Bowden, ``Translating images into maps,'' in \emph{2022 International conference on robotics and automation (ICRA)}.\hskip 1em plus 0.5em minus 0.4em\relax IEEE, 2022, pp. 9200--9206.

\bibitem{song2023fedbevt}
R.~Song, R.~Xu, A.~Festag, J.~Ma, and A.~Knoll, ``Fedbevt: Federated learning bird's eye view perception transformer in road traffic systems,'' \emph{IEEE Transactions on Intelligent Vehicles}, 2023.

\bibitem{pan2023baeformer}
C.~Pan, Y.~He, J.~Peng, Q.~Zhang, W.~Sui, and Z.~Zhang, ``Baeformer: Bi-directional and early interaction transformers for bird's eye view semantic segmentation,'' in \emph{Proceedings of the IEEE/CVF Conference on Computer Vision and Pattern Recognition}, 2023, pp. 9590--9599.

\bibitem{liu2023petrv2}
Y.~Liu, J.~Yan, F.~Jia, S.~Li, A.~Gao, T.~Wang, and X.~Zhang, ``Petrv2: A unified framework for 3d perception from multi-camera images,'' in \emph{Proceedings of the IEEE/CVF International Conference on Computer Vision}, 2023, pp. 3262--3272.

\bibitem{chambon2024pointbev}
L.~Chambon, E.~Zablocki, M.~Chen, F.~Bartoccioni, P.~P{\'e}rez, and M.~Cord, ``Pointbev: A sparse approach for bev predictions,'' in \emph{Proceedings of the IEEE/CVF Conference on Computer Vision and Pattern Recognition}, 2024, pp. 15\,195--15\,204.

\bibitem{smith2019super}
L.~N. Smith and N.~Topin, ``Super-convergence: Very fast training of neural networks using large learning rates,'' in \emph{Artificial intelligence and machine learning for multi-domain operations applications}, vol. 11006.\hskip 1em plus 0.5em minus 0.4em\relax SPIE, 2019, pp. 369--386.

\bibitem{bartoccioni2023lara}
F.~Bartoccioni, {\'E}.~Zablocki, A.~Bursuc, P.~P{\'e}rez, M.~Cord, and K.~Alahari, ``Lara: Latents and rays for multi-camera bird’s-eye-view semantic segmentation,'' in \emph{Conference on Robot Learning}.\hskip 1em plus 0.5em minus 0.4em\relax PMLR, 2023, pp. 1663--1672.

\end{thebibliography}
\newpage

\end{document}